\documentclass{ieeeaccess}
\usepackage{cite}
\usepackage{amsmath,amssymb,amsfonts}
\usepackage{wrapfig}
\usepackage{algorithmic}
\usepackage{algorithm}
\usepackage{multirow}
\usepackage{dsfont}
\usepackage{array}
\usepackage{graphicx}
\usepackage{textcomp}
\usepackage{hyperref}
\usepackage{url}
\def\BibTeX{{\rm B\kern-.05em{\sc i\kern-.025em b}\kern-.08em
    T\kern-.1667em\lower.7ex\hbox{E}\kern-.125emX}}
\begin{document}
\newcolumntype{P}[1]{>{\centering\arraybackslash}p{#1\linewidth}}
\newcolumntype{L}[1]{>{\raggedright\arraybackslash}p{#1\linewidth}}

\history{Date of publication xxxx 00, 0000, date of current version xxxx 00, 0000.}
\doi{xx.xxxx/ACCESS.2022.DOI}

\title{Deep Reinforcement Learning for System-on-Chip: Myths and Realities}
\author{\uppercase{Tegg Taekyong Sung}\authorrefmark{1} and
\uppercase{Bo Ryu\authorrefmark{2}}}
\address[1]{EpiSys Science, Inc., Poway, CA 92064, USA (e-mail: tegg@episyscience.com)}
\address[2]{EpiSys Science, Inc., Poway, CA 92064, USA (e-mail: boryu@episyscience.com)}

\corresp{Corresponding author: Bo Ryu (e-mail: boryu@episyscience.com)\\
This work has been submitted to the IEEE for possible publication. Copyright may be transferred without notice, after which this version may no longer be accessible.}

\begin{abstract}
Neural schedulers based on deep reinforcement learning (DRL) have shown considerable potential for solving real-world resource allocation problems, as they have demonstrated significant performance gain in the domain of cluster computing. In this paper, we investigate the feasibility of neural schedulers for the domain of System-on-Chip (SoC) resource allocation through extensive experiments and comparison with non-neural, heuristic schedulers. The key finding is three-fold. First, neural schedulers designed for cluster computing domain do not work well for SoC due to i) heterogeneity of SoC computing resources and ii) variable action set caused by randomness in incoming jobs. Second, our novel neural scheduler technique, Eclectic Interaction Matching (EIM), overcomes the above challenges, thus significantly improving the existing neural schedulers. Specifically, we rationalize the underlying reasons behind the performance gain by the EIM-based neural scheduler. Third, we discover that the ratio of the average processing elements (PE) switching delay and the average PE computation time significantly impacts the performance of neural SoC schedulers even with EIM. Consequently, future neural SoC scheduler design must consider this metric as well as its implementation overhead for practical utility.
\end{abstract}


\begin{keywords}
Deep reinforcement learning, heuristic scheduler, neural scheduler, resource allocation, system-on-chip scheduling 
\end{keywords}

\titlepgskip=-15pt

\maketitle

\section{Introduction}\label{sec:introduction}

\PARstart{A}{pproaching} the limit of Moore's Law has spurred tremendous advances in System-on-Chip (SoC) which bestows unprecedented gain in computational and energy efficiency for a wide range of applications through an integrated architecture of general-purpose and specialized processors~\cite{hennessy2019new}. In particular, the domain-specific SoC (DSSoC), a class of heterogeneous chip architecture, empowers exploitation of distinct characteristics of compute different resources (\textit{i.e.}, CPU, GPU, FPGA, accelerator, etc.) for speed maximization and energy efficiency via intelligent resource allocation~\cite{krishnakumar2022design,moazzemi2020runtime,mettler2022fpga}. The primary goal of a DSSoC scheduling policy is to optimally assign a variety of hierarchically structured jobs, derived from many-core platforms executing streaming
applications from wireless communications and radar systems, to heterogeneous resources or processing elements (PEs). Over the years, researchers have demonstrated effective performance for DSSoC with expert-crafted, heuristic rules~\cite{arda2020ds3,mack2021performant}.

\renewcommand{\arraystretch}{1.3}
\begin{table*}[!t]
\centering
\begin{tabular}{ P{0.1}P{0.09}L{0.31}P{0.1}P{0.1}P{0.15} } 
\hline
Algorithm & Application & \centering Approach & Job & Resource & Objective \\
\hline
\multirow{2}{*}{TetriSched~\cite{tumanov2016tetrisched}} & \multirow{2}{*}{Cluster} & Estimates job run-time heuristically and plans for placement options & MapReduce jobs & Heterogeneous clusters & Minimization of errors in job execution timing  \\
\multirow{2}{*}{Decima~\cite{mao2018learning}} & \multirow{2}{*}{Cluster} & Allocates a quantity of resources to ready tasks using graph-structured information  & \multirow{2}{*}{Spark jobs} & Homogeneous clusters & Minimization in avg. job completion time  \\ 
\multirow{2}{*}{Gandiva~\cite{xiao2018gandiva}} & \multirow{2}{*}{Cluster} & Exploits job predictability to time-slice resources efficiently across multiple jobs  & \multirow{2}{*}{DLT jobs} & Heterogeneous clusters & Improvement on cluster utilization \\
\multirow{2}{*}{Tiresias~\cite{gu2019tiresias}} & \multirow{2}{*}{Cluster} & Assigns job priority using Gittins index to schedule distributed jobs & \multirow{2}{*}{DLT jobs} & Homogeneous clusters & Minimization in avg. job completion time\\
\multirow{2}{*}{Themis~\cite{mahajan2020themis}} & \multirow{2}{*}{Cluster} & Uses a two-level architecture to capture placement sensitivity and ensure efficiency & \multirow{2}{*}{DLT jobs} & Heterogeneous clusters & Improvement on cluster utilization \\
\multirow{2}{*}{Allox~\cite{le2020allox}} & \multirow{2}{*}{Cluster} & Transforms the scheduling problem into a min-cost bipartite matching problem & \multirow{2}{*}{DLT jobs} & Heterogeneous clusters & Minimization in avg. job completion time \\
\multirow{2}{*}{Gavel~\cite{narayanan2020heterogeneity}} & \multirow{2}{*}{Cluster} & Generalizes existing scheduling policies by expressing them as optimization problems & \multirow{2}{*}{DLT jobs} & Heterogeneous clusters & Minimization in avg. job completion time. \\
\multirow{2}{*}{DeepRM~\cite{mao2016resource}} & \multirow{2}{*}{Cluster} & Includes backlog information on remaining jobs; train the agent using REINFORCE & \multirow{2}{*}{Cluster jobs} & \multirow{2}{*}{Single cluster} & Minimization in avg. slowdown\\
\multirow{2}{*}{SCARL~\cite{cheong2019scarl}} & \multirow{2}{*}{Cluster} & Employs attentive embedding and schedules tasks using factorization of action & \multirow{2}{*}{Cluster jobs} & Heterogeneous clusters & Minimization in avg. slowdown \\ 
\multirow{2}{*}{DRM~\cite{sung2019neural}} & \multirow{2}{*}{SoC} & Iteratively maps tasks to resources and updates the agent using REINFORCE algorithm & Single synthetic job & Heterogeneous resources & Minimization in job completion time \\
\multirow{2}{*}{DeepSoCS~\cite{sung2020deepsocs}} & \multirow{2}{*}{SoC} & Re-arranges task orders using graph-structured information and greedily maps tasks to resources & Synthetic and SoC jobs & Heterogeneous resources & Minimization in avg. latency \\
\multirow{2}{*}{SoCRATES}~\cite{sung2021socrates} & \multirow{2}{*}{SoC} & Iteratively maps tasks to resources and aligns post-processed returns to corresponding tasks & Synthetic and SoC jobs & Heterogeneous resources & Minimization in avg. latency \\
\hline
\end{tabular}
\caption{Design features of cluster and DSSoC scheduling approaches (DLT: Deep Learning Training).}
\label{table:sched-comp}
\end{table*}

While heuristic schedulers have been dominant in a wide range of domains for resource allocation, recent effort on scheduling algorithm development started undergoing a paradigm shift toward neural approaches as they demonstrated state-of-the-art performance in complex resource management domains~\cite{holt2020novel,goksoy2021dynamic}. In particular, recent successes in applying deep reinforcement learning (DRL) for scheduling heterogeneous (cloud) cluster resources~\cite{mao2018learning,cheong2019scarl} have further motivated applying similar DRL approaches for task scheduling on DSSoC, obtaining noticeable performance gains over well-known heuristic schedulers under certain operational conditions~\cite{sung2019neural,sung2020deepsocs,sung2021socrates}. Through extensive experimentation with both DRL and heuristic schedulers under extremely wide ranges of DSSoC scenarios, we present an in-depth comparative analysis between neural schedulers and their heuristic counterparts for the DSSoC domain. The key contribution of our research is that the high performance of DRL schedulers previously observed in both cloud cluster and DSSoC domains is found to be highly sensitive to the ratio of the average PE switching delay and the average PE computation time. Specifically, when this ratio is close to one, neural schedulers tend to outperform their heuristic counterparts under various operational scenarios. On the other hand, when the ratio is much less than one and subject to other operational conditions, the anticipated high performance of neural schedulers does not materialize. We attribute this to two major factors: (i) heterogeneity of SoC computing resources; (ii) variable action set caused by randomness in incoming jobs. When combined, they exacerbate the problem of \textit{delayed reward} because the accumulated rewards are likely to disrupt the backpropagation-based optimization method. With this finding, we present a realistic avenue for future DRL-based resource scheduler design.

\section{Related Work}\label{sec:related}

Design of high-performance SoC resource schedulers has been active for many years~\cite{arda2020ds3,mack2021performant}. Scheduling algorithms are mostly heuristic in nature with specific optimization goals. Examples include First Come First Served (FCFS), Earliest Task First (ETF)~\cite{blythe2005task}, Minimum Execution Time (MET)~\cite{buttazzo2011hard}, and Hierarchical Earliest First Time (HEFT)~\cite{topcuoglu1999task}. While both MET and STF schedule tasks to PEs which take the shortest amount of execution time, HEFT schedules tasks by considering both task computation time and data transmission delays. A real-time heterogeneity-aware scheduler HetSched~\cite{amarnath2021heterogeneity} with task- and meta-scheduling components having multiple static DAG-represented jobs as input is built for autonomous vehicle applications. A new pruning Monte-Carlo Tree Search (MCTS)-based algorithm~\cite{kung2022improved} has been applied for workflow scheduling. It has improved performance in makespan over the heuristics, Improved Predict Priority Task Scheduling (IPPTS)~\cite{djigal2020ippts}, and a meta-heuristic Genetic Algorithm approach~\cite{keshanchi2017improved}. However, much of the gain depends on the specific heuristics and the nature of job configurations.

Cluster resource management for cloud computing (\textit{e.g.}, YARN~\cite{vavilapalli2013apache} or Kubernetes~\cite{burns2016borg}) is another orthogonal approach in resource allocation. It is primarily designed to schedule big-data, time-persistent jobs (\textit{i.e.}, MapReduce~\cite{dean2008mapreduce} or Deep Learning Training (DLT) jobs\footnote{A neural network represents a job, and each operation, such as matrix multiplication or nonlinear function, acts as tasks.}). A list of scheduling approaches along with their design features is summarized in Table~\ref{table:sched-comp}. Themis~\cite{mahajan2020themis} and Tiresias~\cite{gu2019tiresias} allocate tasks from distributed DLT jobs to homogeneous clusters using two-dimensional scheduling algorithm. Gandiva~\cite{xiao2018gandiva} schedules a set of heterogeneous DLT jobs to a fixed set of GPU clusters. It allows preemption on jobs to share overload jobs to available resources' spaces. AlloX~\cite{le2020allox} transforms a heterogeneous resources scheduling problem into a min-cost bipartite matching problem in order to provide performance optimization and fairness to users in Kubernetes. TetriSched~\cite{tumanov2016tetrisched} estimates job run-time heuristically for placement options. Gavel~\cite{narayanan2020heterogeneity} transforms existing scheduling policies to heterogeneity-aware optimization problems for generalization and improves the diversity of policy objectives. Such cluster schedulers enhance run-time performance by exploiting the simulation characteristics.

Neural schedulers have begun to surpass hand-crafted algorithms and show a significant performance gain in the cluster scheduling problem. DeepRM, the first DRL-based cluster scheduler reported in the literature, shows significant reduction in job slowdown\footnote{This metric represents a relative value of actual job duration and ideal job duration.} over heuristics~\cite{mao2016resource}. In comparison to DeepRM, Decima~\cite{mao2018learning} proposes an end-to-end neural scheduler for more realistic cluster environment with hierarchical cluster jobs. It extracts hierarchical job information with graph neural networks (GNNs)~\cite{gilmer2017neural} and decides how many resources to execute each task. Decima addresses varying action selection caused by the hierarchical jobs using placeholder implementation~\cite{abadi2016tensorflow}, but it considers homogeneous clusters. SCARL~\cite{cheong2019scarl} aims to schedule jobs to heterogeneous resources by exploiting attentive embedding~\cite{vaswani2017attention} in policy networks. However, SCARL is not able to schedule non-hierarchical jobs, which differs from Decima, and is not applicable to the realistic environment~\cite{mao2018learning}. Spear~\cite{hu2019spear} applies MCTS to plan the task scheduling with a DRL model for guidance in the expansion and rollout steps in MCTS. 

Building on the success of neural schedulers for the cluster environment as described above, novel neural approaches have been proposed for the domain of SoC. Deep Resource Management (DRM)~\cite{sung2019neural} is considered the first DRL-based SoC scheduler that schedules hierarchical jobs to heterogeneous resources in the scenario with a single synthetic job. DeepSoCS~\cite{sung2020deepsocs}, adapted from Decima, is proposed for handling more realistic SoC scenarios where multiple numbers of both synthetic and real-world SoC jobs are continuously generated. It is a hybrid approach that rearranges the tasks using the graph-structured information extracted by GNNs and maps them to resources using a heuristic algorithm. However, the performance gain achieved by DeepSoCS depends on operational conditions, as it is inherently imitating the expert policy with an exhaustive search employed by heuristic schedulers. In order to explore the feasibility of an end-to-end neural SoC scheduler with the goal of achieving significant performance gain over heuristic schedulers, the authors proposed SoCRATES~\cite{sung2021socrates} with a novel technique of Eclectic Interaction Matching (EIM). EIM remedies the concurrency problem in receiving observation and reward gains by matching the time-varying interaction and simulation time steps. Consequently, SoCRATES achieves considerable enhancement in performance over prior neural schedulers~\cite{sung2019neural,sung2020deepsocs}. In this paper, we present key insights into how such performance gain is achieved by SoCRATES through extensive comparative experimentation.

\section{Motivation}\label{sec:motivation}
Despite significant performance gains demonstrated by neural schedulers for cluster computing management, they generally suffer from limited extensibility. For example, prior cluster schedulers address non-hierarchical workloads~\cite{le2020allox,xiao2018gandiva} and homogeneous resources~\cite{mao2016resource,mao2018learning} that cannot fully exhibit SoC resource allocation. Although a series of research in the cluster application employs heterogeneous machines~\cite{tumanov2016tetrisched,xiao2018gandiva,mahajan2020themis,le2020allox}, their complexity is relatively simpler than DSSoC. Schedulers in cluster applications allocate jobs to a set of CPUs or GPUs with different performances, whereas schedulers in DSSoC applications map a range of domain-specific jobs to various types of PEs, e.g., CPUs, GPUs, accelerators, memory, each with different performance and supported functionalities. Cluster schedulers decide how many clusters to execute incoming tasks, whereas SoC schedulers map which SoC computing resource to an incoming task. Hence, the scheduler must be aware of unsupported action for an individual task. Hence, directly applying neural schedulers to SoC is non-trivial due to the disparities in the environment properties, such as the structures of jobs/resources and scheduling mechanisms.

In contrast, heuristic scheduling algorithms in the domain of SoC steadily show state-of-the-art performance. We discovered that their significant performance gains come from rescheduling task assignments with exhaustive searches, such as PE availability checks or gaps between consecutive task assignments (see Section~\ref{sec:eval:rule-based} for more details). However, such rule-based algorithms generally have limited robust performance. For instance, heuristic schedulers are vulnerable to system perturbation from external forces in the setting of a single job execution~\cite{sung2019neural}. Based on these robust and significant performance gains in the domain of cluster computing, we are interested in extending these neural schedulers to SoC. While neural algorithms generally adapt to dynamic system changes and have robust performance~\cite{sutton2018reinforcement}, subsequent works have motivated and developed in a more complicated and practical scenario with continuous job injection~\cite{sung2020deepsocs,sung2021socrates}. In this paper, we investigate the challenging standpoints for designing DRL scheduling policy in the domain of SoC. With the recently introduced EIM technique overcoming such challenges, we rationalize the underlying reasons behind the performance improvement in existing neural schedulers by examining PE usages and action designs. Furthermore, we investigate which operation condition impacts the performance of neural SoC schedulers with EIM. To the end, the questions we want to consider in this paper are the following:
\begin{itemize}
    \item What is the main difference between SoC and other domains?
    \item How does the neural scheduling policy effort change in SoC domain?
    \item Under which operational conditions do neural schedulers perform/cannot perform well?
    \item How does EIM technique improve the performance of neural schedulers?
    \item What are the strengths/weaknesses of neural scheduler?
\end{itemize}

\section{Background and system model}\label{sec:background}

A large body of research in scheduling exists for a broad range of domains. Cluster management in datacenters allocates Spark or DLT jobs to a set of CPU and GPU machines. This paper contributes to the domain of heterogeneous DSSoC and its emulator in the form of high-fidelity Domain-Specific SoC Simulator (DS3)~\cite{arda2020ds3,sung2021gymds3}. DS3, which supports a heterogeneous SoC computing platform Odroid-XU3~\cite{ODROID-XU3}, enables the allocation of a set of communication or radar jobs to various types of resources, such as general-purpose cores, hardware accelerators, and memory. The ARM heterogeneous big.LITTLE architecture of the cores enables performance-oriented and energy-efficiency runs (the big cores of 2.1Ghz Cortex-A15 are performance-oriented, the LITTLE cores of 1.5GHz Cortex-A7 are energy-efficient). DS3 integrates system-level design features for hierarchical jobs and heterogeneous resources. The job and resource profiles are given as a list specifying properties. The system parses them and generates workloads and PEs using the job and resource models.

\subsection{System-on-chip simulation}\label{sec:background:sim}

\begin{figure}[!t]
\centering
\includegraphics[width=0.478\textwidth]{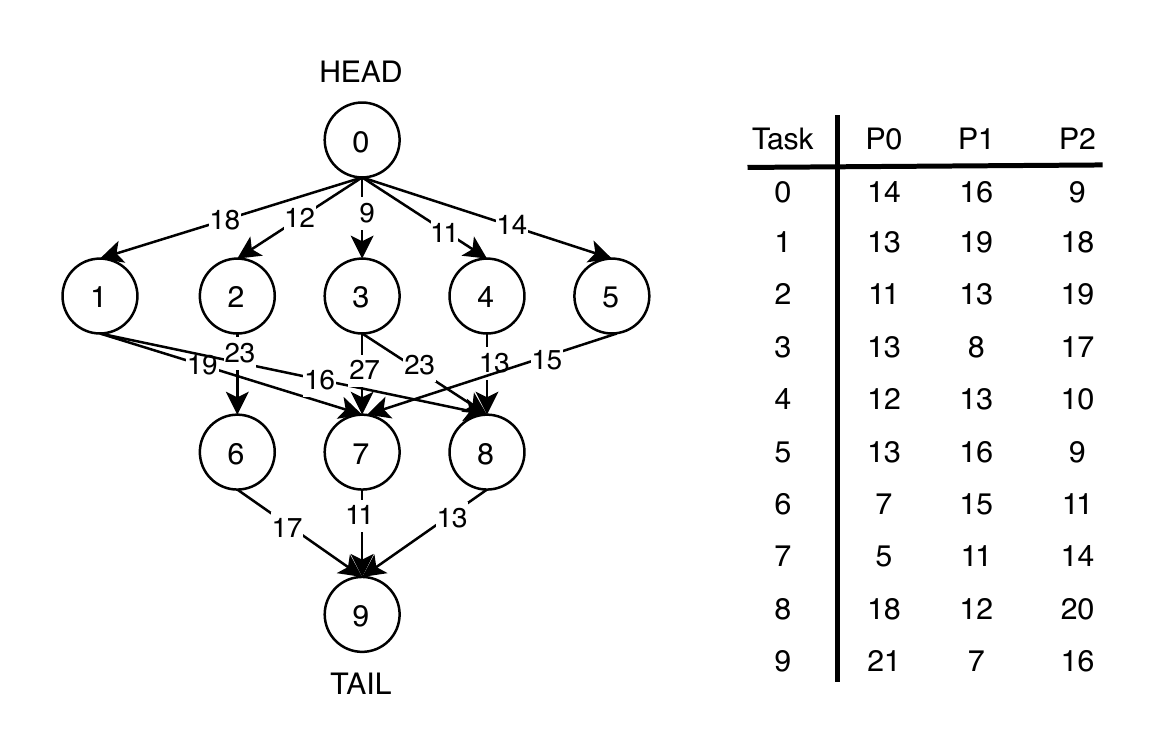}
\caption{\label{fig:job_dag} An illustration of a set of synthetic job and resource profiles. The diagram on the left depicts a job DAG, where a note represents a task by its ID and the edge represents data transmission delay by its weights. The table on the right shows a set of heterogeneous PEs with different computation time for each task.}
\end{figure}

\subsubsection{Job Model}\label{sec:background:sim:job}
We define a job as a collection of interleaved tasks. Jobs in DS3 implement real-world applications of wireless communication and radar processing. The tasks represent operations, such as waveform generator, Fast Fourier Transform, vector multiplication, or decoder~\cite{arda2020ds3}. A \textit{job structure} is in the form of a directed acyclic graph (DAG), illustrated in Fig.~\ref{fig:job_dag}~\cite{topcuoglu1999task}. A job is denoted as $G=(N,E)$, where $N$ is a set of nodes and $E$ is a set of edges. Each node $n_i \in N$ represents a heterogeneous task in the job, and each directed edge $e_{i,j}\in E$ connects node $n_i$ to node $n_j$. We interchangeably use the term ``task'' and ``node'' unless there is confusion. The edge encodes task dependency. To be specific, if there exists edge $e_{i,j}$, task $n_j$ can start execution only after task $n_i$ finishes. Here, we call node $n_i$ a parent of node $n_j$, and $n_j$ a child of $n_i$. We define a set of parents of $n_j$ (predecessors) by $pred(n_j)$ and a set of child of $n_i$ (successors) by $succ(n_i)$. A node may have multiple parents or children, and nodes can be simultaneously executed. Each edge $e_{i,j}$ has a weight $w_{i,j}$ that represents data transmission delay between $n_i$ and $n_j$. This delay is added to the task duration when the scheduler selects a different PE to task $i$ from task $j$. The labels with HEAD and TAIL refer to the root parent node and the terminal leaf node, respectively. Assume a job $G$ has $v$ tasks, $N=\{n_1,\dots,n_v\}$, then the job is considered complete when all tasks in $N$ have been completed. Here, $n_1$ is HEAD node and $n_v$ is TAIL node. According to the job model aforementioned, multiple jobs are generated. Each job is generated based on the following parameters~\cite{topcuoglu1999task}:
\begin{enumerate}
    \item $v$: the number of tasks in the directed acyclic graph
    \item $\alpha$: the shape parameter of the graph. $\alpha$ controls the width and depth of a graph structure. We sample the average width of each level in a graph from a normal distribution with a mean of $\sqrt{v} \times \alpha$. The depth of a graph is equal to the $\frac{\sqrt{v}}{\alpha}$ (see Appendix~\ref{sec:append:job_construct} for details on job DAG construction). If $\alpha \gg 1.0$, a shallow but wide graph is generated; if $\alpha \ll 1.0$, a deep but narrow graph is generated.
    \item $\nu$: the average value of communication delay. The weight of $e_{i,j}$, representing communication delay, is set to $\max(1,\lfloor |w| \rfloor)$, where $w \sim \mathcal{N}(\nu, 0)$.
    \item $CCR$: the communication-to-computation ratio. We calculate an average communication cost by the sum of the scheduled PE bandwidth and the weights of edges between the current task and the previous task. An average computation cost is defined in the SoC job profile. If the $CCR$ value in a DAG is high, the job is a communication-intensive workload. Conversely, if the $CCR$ value is low, the job is a computation-intensive workload.
    \item $d_{in}$: an average value of in-degree of nodes
    \item $d_{out}$: an average value of out-degree of nodes
\end{enumerate}

\subsubsection{Resource Model}\label{sec:background:sim:res}
Resource profile defines the characteristics of PEs, and  each PE is defined with a set of different, fixed supported tasks and operating performance points (OPP). OPP is a utilization set for a tuple of power consumption and task run-time frequency. OPP for PE $q$, for instance, can be defined by a set of voltage-frequency pairs, $\mathcal{OPP}_q=\{(V^q_1,f^q_1),\dots,(V^q_O,f^q_O)\}$ where $O$ is the number of operating points. Once the  frequency parameter is given, the resource model creates the corresponding PE. Since PE running with high frequency, generally, executes tasks faster but consumes more power and energy, a trade-off exists between run-time performance and energy efficiency. Moreover, each PEs has a bandwidth that contribute to the communication delay when the simulator switches over PEs during task execution.

\subsubsection{Objective}\label{sec:background:sim:objective}
DS3 is heavily shaped by the peculiarities of the SoC domain. DS3 comes with real-world reference applications from wireless communications and radar processing domains. Each supported workload consists of various operations (\textit{i.e.}, tasks), which require a short amount of duration. The run-time overhead of each task includes the task duration and data transmission delay. The allocation of different processors at the same time for task $n_i$ and its parent task set $\{n_j\}=pred(n_i)$ would incur the data transmission delay. Let task $n_i$ is mapped to PE $P_i$ and its task computation time with operating frequency $f^i_o$ by $comp(n_i | P_i,f^i_o)$. Then, the overall task duration is equated by
\begin{equation}
    exec(n_i) = \mu \cdot comp(n_i|P_i,f^i_o) + delay(n_i),
\label{eq:exec_time}
\end{equation}
\noindent where $\mu$ indicates a scaling parameter for extending the task execution time. On the right-hand side, the first term is task computation time on a PE, and the second term is data communication delay, given by:
\begin{equation}
    delay(n_i) = \max_{n_j \in pred(n_i)} \frac{w_{i,j}}{B(P_i,P_j)},
\label{eq:comm_delay}
\end{equation}
\noindent where $w_{i,j}$ is the weight of edges between task $i$ and task $j$, and $B(P_i,P_j)$ is the PE bandwidth from $P_i$ to $P_j$. The self-loop bandwidth of the same processors is assumed to be negligible, $B(P_i,P_i)=0$. Due to the communication delay, frequent resource switching leads to an increasing loss in task completion time. The objective to optimize, the \textit{average latency} minimization, is given by
\begin{equation}
    L = \frac{\sum_{G\in\mathcal{G}_{comp}}\sum_{i\in |G|}exec(n_i)}{|\mathcal{G}_{comp}|},
\label{eq:latency}
\end{equation}
\noindent where $\mathcal{G}_{comp}$ is a set of completed job DAGs, and $|G|$ is the number of tasks in the job $G$. 
\begin{figure}[!t]
\centering
\includegraphics[width=0.5\textwidth]{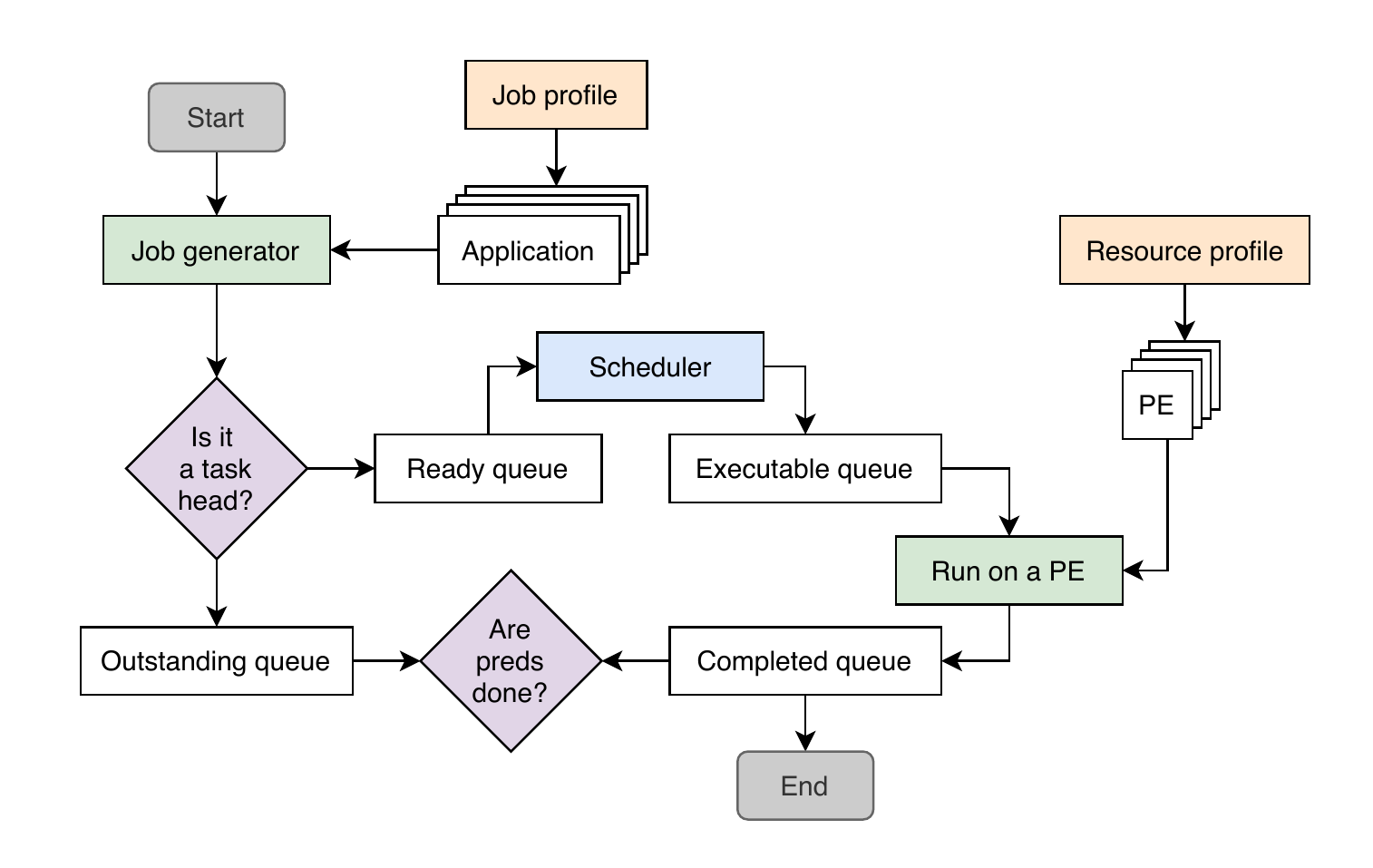}
\caption{\label{fig:ds3_workflow} An overview of DS3 workflow. At initialization, a set of workloads and PEs are generated for given job and resource profiles. The job generator continuously generates multiple jobs using the set of workloads and distributes them to the task queues. The scheduler takes any tasks in the ready queue and maps each task to one of the PEs. If the PE is idle, it starts task execution. The task dependency graph prescribes which next task to move onto the ready queue after the completion of its predecessors.}
\end{figure}

Previous work~\cite{topcuoglu1999task} introduced additional evaluation metrics of run-time overhead for a single completed job: Schedule Length Ratio (SLR) and Speedup. The SLR and Speedup metrics are given by
\begin{equation}
    SLR=\frac{makespan}{\sum_{n_i\in CP_{MIN}}\min_{p_j\in Q}\{w_{i,j}\}}
\label{eq:slr}
\end{equation}
\begin{equation}
   Speedup=\frac{\min_{p_j\in Q}\Big\{\sum_{n_i\in v}w_{i,j}\Big\}}{makespan},
\label{eq:speedup}
\end{equation}
\noindent where the denominator of $SLR_k$ represents the ideal lower limit time for scheduling for the job DAG. $CP_{MIN}$ is the minimal critical path of job DAG, and $Q$ is the number of PEs. The nominator of $Speedup$ represents the overall task computation time when each of the $v$ tasks in a job DAG is scheduled onto the same processor. This indicates the ability of the algorithm to schedule tasks to explore parallel performance. The lower $SLR$ and the higher $Speedup$, the more optimal scheduling performance.

Since this paper seeks to evaluate performance over multiple jobs, we average out $SLR$ and $Speedup$ over the entire completed jobs per simulation length. Let a set of completed jobs by $\{G_i\}_{i=1}^{|\mathcal{G}_{comp}|}$, each of the completed jobs corresponds to the set of generated workloads at DS3 initialization. Since heterogeneous jobs are generated, each job likely has a different minimal critical path and parallel performance with the same processors. The average $SLR$ and the average $Speedup$ are given by
\begin{equation}
    \overline{SLR} = \frac{\sum^{|\mathcal{G}_{comp}|}_{k=1}SLR_k}{|\mathcal{G}_{comp}|}
\end{equation}
\begin{equation}
   \overline{Speedup} = \frac{\sum^{|\mathcal{G}_{comp}|}_{k=1}Speedup_k}{|\mathcal{G}_{comp}|}.
\end{equation}

\begin{algorithm}[!t]
\caption{DS3 Environment}
\begin{algorithmic}[1]
\STATE {\bfseries Input:} job inter-arrival rate $scale$, clock signal $clk$, maximum simulation length $CLK$, job model $M_J$, resource model $M_R$, job capacity $C$, job queue $Q_{job}$, ready task queue $Q_{ready}$, job profile \texttt{job}, resource profile \texttt{resource}, $W$ number of jobs, $Q$ number of PEs
\STATE {\bfseries Output:} average latency $L$
\FOR{each episode}
\STATE $clk \leftarrow 0$
\STATE $\{G_i\}_{i=1:W} \leftarrow M_J(\texttt{job})$
\STATE $\{P_i\}_{i=1:Q} \leftarrow M_R(\texttt{resource})$
\REPEAT
\STATE \textcolor[rgb]{0.5,0.5,0.5}{\# Generate jobs}
\IF{$|Q_{job}| < C$}
\STATE $clk_{inj} \sim Exp(scale)$
\STATE $Q_{job} \leftarrow G$ at $clk_{inj}$
\ENDIF
\FOR{each task $i$ in $Q_{ready}$}
\STATE \textcolor[rgb]{0.5,0.5,0.5}{\# Schedule tasks in ready list to PE}
\ENDFOR
\IF{$P$ is idle}
\STATE \textcolor[rgb]{0.5,0.5,0.5}{\# PE execution}
\STATE start $P$ execution corresponding to the scheduled tasks
\ENDIF
\STATE $clk \leftarrow clk + 1$
\UNTIL{$clk = CLK$}
\STATE Compute $L$ using~\eqref{eq:latency}
\ENDFOR
\end{algorithmic}
\label{alg:ds3env_pseudocode}
\end{algorithm}

\begin{table*}[!t]
\centering
\begin{tabular}{ cccc }
& \multicolumn{2}{c}{DS3} & \multirow{2}{*}{Spark} \\
& synthetic & real-world & \\
\hline
\multicolumn{1}{l}{\textbf{Job characteristic}} & & \multicolumn{1}{l}{} \\
\multicolumn{1}{l}{Number of tasks} & 10 & 27 & 1137.6 \\
\multicolumn{1}{l}{Data transmission delays} & $16.6 \pm 5.0$ & $3.38 \pm 2.4$ & $2000$ \\
\multicolumn{1}{l}{Average job DAG level} & $4$ & $7$ & $5.8 \pm 2.6$ \\
\multicolumn{1}{l}{Number of types} & 1 & 1 & $154$ \\
\multicolumn{1}{l}{Average job arrival time} & \multicolumn{2}{c}{$25$} & $25000$ \\
\multicolumn{1}{l}{Job duration} & \multicolumn{2}{c}{Varied} & $1127.3 \pm 441.9$ \\
\hline
\multicolumn{1}{l}{\textbf{Resource characteristic}} & & \multicolumn{1}{l}{} \\
\multicolumn{1}{l}{Structure} & \multicolumn{2}{c}{Heterogeneous} & Homogeneous \\
\multicolumn{1}{l}{Task computation time} & $13.3 \pm 4.1$ & $40.0 \pm 83.6$ & -  \\
\multicolumn{1}{l}{Number of type} & 4 & 17 & $1$ \\
\hline
\end{tabular}
\caption{\label{tbl:diff_envs} A comparison between DS3 and Spark properties. Due to the differences in applicability, DS3 and Spark have different characteristics of job and resource. On DS3, a representative real-world profile, WiFi-TX, is included. Note that the shape parameter controls the diversity in the number of job types and their average graph levels, $\alpha$.}
\end{table*}

The DS3 workflow is given in Algorithm~\ref{alg:ds3env_pseudocode} and Fig.~\ref{fig:ds3_workflow}. After initialization, DS3 continuously generates indefinite hierarchical-structured workloads with respect to the job model at stochastic job inter-arrival rates. While the number of injected workloads is below the job capacity $C$, the job generator injects a mix of multiple instances of the workloads in a stream fashion, $\mathcal{G}=\{G_1,\dots,G_W\}$, where $W \leq C$. The workloads are generated at every $clk_{inj}$, where $clk_{inj} \sim Exp(scale)$, where $scale$ is the mean of job inter-arrival rate. A large value of job inter-arrival leads to high frequency in job injection. Then, DS3 loads a set of tasks that have no dependency onto the ready queue, otherwise onto the outstanding queue. Each ready task, which is derived dynamically based on the prior task scheduling, is ready to be scheduled by PEs using a scheduling policy. The task then moves to the executable queue, and the corresponding PE, if idle, starts executing the task. The tasks are non-preemptive, as DS3 runs in a non-preemptive setting during task execution. The job generator, distributed PEs, and simulation kernel, all of which are executed in parallel, share the same clock signal.

\subsection{Simulation analysis}\label{sec:background:analysis}
Elucidating the distinction in simulation behaviors, we compare DS3 against Spark~\cite{mao2018learning}, one of the representative realistic simulations for cluster applications. Both simulations support the scheduling of multiple graph-structured jobs, but differ greatly in the mechanism of resource allocation in their domains. We list the job and resource characteristics after normalization in Table~\ref{tbl:diff_envs}.

The scheduling policy in Spark decides on how many resource machines to allocate for the ready tasks with respect to the given job profile. DS3 scheduling policy, on the other hand, decides which PE to execute the ready tasks, and the task run-time is solely dependent on the selected PE performance. After task completion, Spark applies a static moving delay, whereas DS3 applies a dynamic data transmission delay. Cluster jobs generally consist of numerous tasks and last for a long time. Spark, for instance, supports 154 types of jobs with approximately 5.8 levels (DAG depth). Alternatively, DSSoC jobs are executed in a short range of duration. DS3 provides several types of real-world job profiles, but this paper focuses on one real-world job, WiFi-TX, and one synthetic job. These jobs have 4 and 7 levels, respectively. Endowing with heterogeneous resources, DS3 has 4 PEs on a synthetic profile and 17 PEs on a real-world profile. Each has a different run-time performance for tasks and different supported functionalities. In that sense, an individual scheduling task must check whether it can be executed in PE. Regarding $CCR$, the synthetic profile has a similar range of computation and communication costs. In contrast, the real-world profile is chain-structured and compute-intensive. That being said, the communication time for the synthetic job has at most 22x larger than that for the real-world job, and the task computation time for real-world resources is at most 13.4x larger than that for synthetic resources. In practice, we modify the job characteristics using shaping parameters $\alpha$, $\mu$, and $\nu$ to grant more variability. (see Section~\ref{sec:background:sim:job} for details on the parameter description). The difference in the mechanisms in two different domains limits the scope of the applicability of each scheduling algorithm, and the extent or range of run-time is largely different.

\begin{figure}[!t]
\centering
\includegraphics[width=0.478\textwidth]{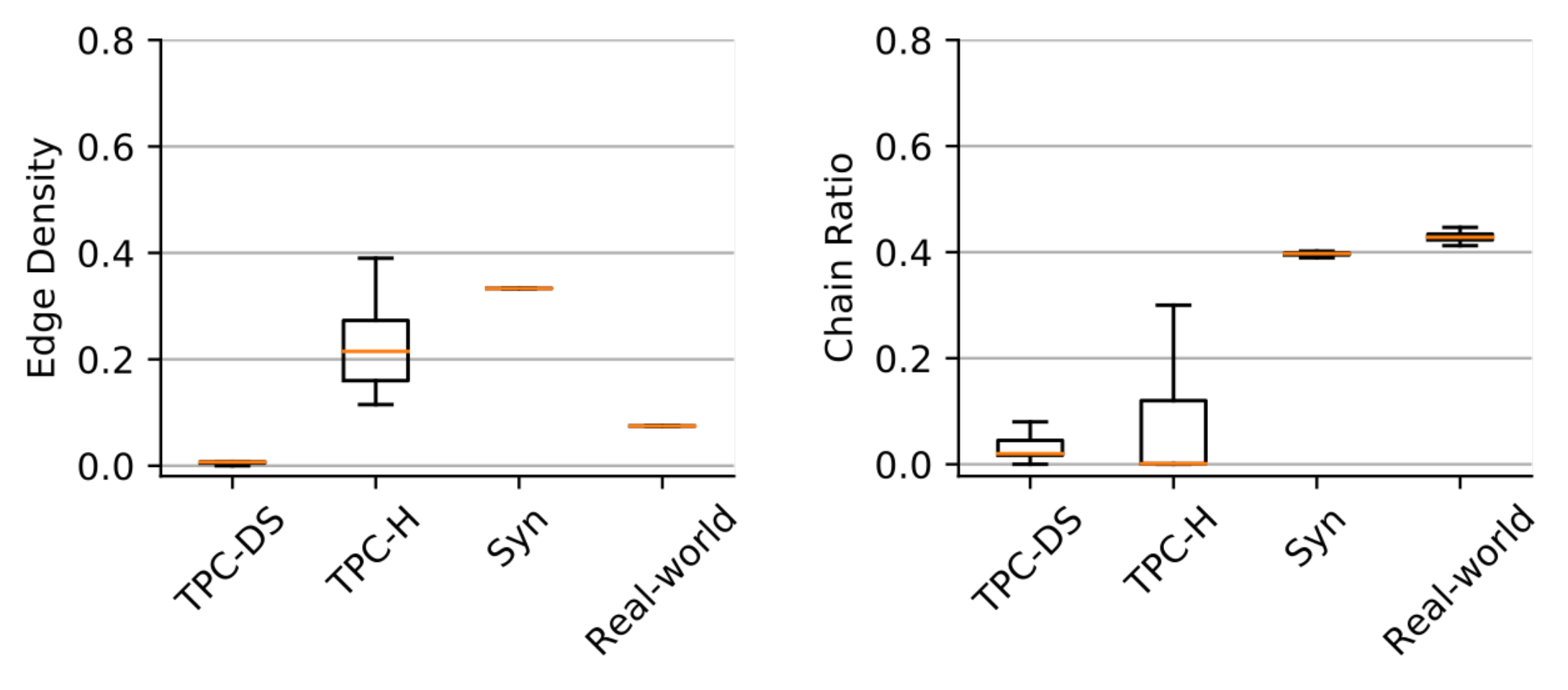}
\caption{\label{fig:edge_chain} The edge density and chain ratio of cluster and SoC workloads. The results of TPC-DS and TPC-H are reproduced by referring to~\cite{tian2019characterizing}.}
\end{figure}

Additional metrics are given here for hierarchical job DAGs~\cite{tian2019characterizing}: (i) The edge density measures the sparsity of job DAGs. The density computed by $\frac{2E}{V(V-1)}$, where $E$ denotes the number of edges, $V$ the number of vertices (tasks), and $V(V-1)$ the possible maximum number of edges for a job DAG—the higher density results in denser job DAG with more complexity. (ii) The chain ratio measures the prevalence of chained tasks. The rate is computed by $\frac{C}{V}$, where $C$ denotes the number of chained tasks that have an exact one child and one parent. Fig.~\ref{fig:edge_chain} reports that the synthetic job DAG is relatively sparse, and SoC jobs have a larger number of chains than cluster jobs.

\begin{figure}[!t] 
\centering
\includegraphics[width=0.42\textwidth]{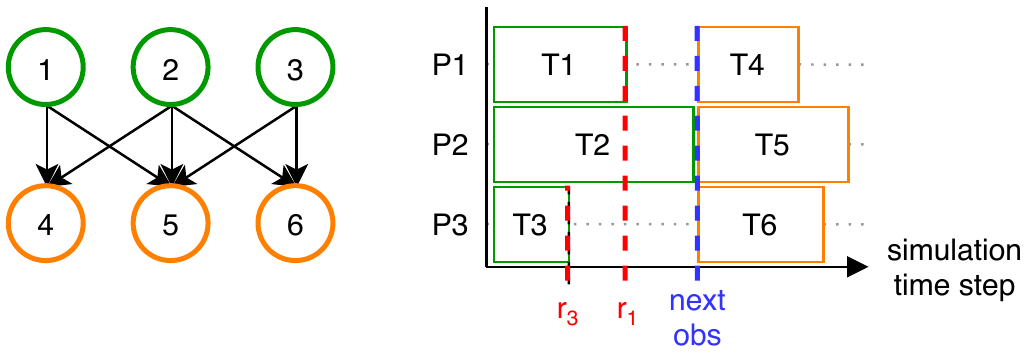}
\caption{\label{fig:irregular_interaction} An illustration of irregular interactions. Although Tasks 1 and 3 have been completed earlier, the next Tasks 4, 5, and 6 are scheduled after Task 2 has been completed. As a result, the reward gains for scheduling decisions for Tasks 1 and 3 are truncated due to the task dependencies.}
\end{figure}

A major difficulty in designing a DS3 scheduler is that the number of available actions (scheduling decisions) varies over time due to the mixes of incoming heterogeneous jobs and their different task dependency graphs. Fig.~\ref{fig:irregular_interaction} illustrates an exemplary scenario where tasks 4, 5, and 6 are a child of task 2. Although task 1 and task 3 have been completed earlier, per the dependency graph, the next observation can be received after completing task 2. Then, the immediate reward signals for tasks 1 and 3 are naturally delayed. With heterogeneous PEs- and dependency-graph-induced variations incurring abrupt dynamic run-time of tasks, it becomes an entangled affair in pairing action and its reward to compute returns properly. Indeed, this entanglement of the task dependency graph and heterogeneous resources leads to the misalignment of the order and timing of observation and reward gains. In that sense, the agent has a mismatched reward timestamp with the actual simulation running clock signal. As a result, the interactions become inconsistent, and rewards (returns) will be incorrectly assessed and backpropagated.

Based on the above analysis, DS3 exhibits dynamic, realistic operational behaviors but differs significantly from other domains of resource allocations. Due to task dependencies from mixes in various jobs, the scheduler must address a variable action set (i), shown in the list below. The distributed PE executes each scheduled task accordingly (ii). By combining (i) and (ii), the agent naturally has delayed rewards likely to disrupt DRL optimization. That is the last difficulty, (iii).
\begin{enumerate}
    \item \textbf{Variable action sets}: Mixes in various jobs with different task dependency graphs cause variable action sets. Since the job queue holds multiple heterogeneous jobs, the agent must recognize multiple job graphs and respond to the fact that action sets are irregular. At every scheduling interaction, the agent receives tasks free of dependencies for the given state.
    \item \textbf{Heterogeneous resources}: As heterogeneity in both jobs and SoC computing resources, the DS3 scheduler must consider different task execution times and data transmission delays. The SoC scheduler computes which task to be executed on which PE. Based on the task-PE mappings from a scheduling policy, the average job duration becomes highly unpredictable.
    \item \textbf{Delayed rewards}: A combination of the varying actions caused by randomness in incoming jobs and heterogeneous resources exacerbates the problem of delayed rewards. The accumulated rewards tend to disrupt DRL optimization. With the previous action commitments and a varying number of observations, the returns must match the interaction steps and the actual simulation clock signal.
\end{enumerate}

\subsection{Benchmark scheduler}\label{sec:background:scheduler}

\subsubsection{Rule-based scheduler}\label{sec:background:scheduler:rule}
The task duration depends on task computation time on a PE and communication delay computed by the PE bandwidth and data transmission delay in the job DAGs, as described in~\eqref{eq:exec_time}. Shortest Time First (STF) and Minimum Execution Time (MET)~\cite{braun2001comparison} iteratively schedule ready tasks to the PE that has minimal execution time. After the schedules, MET checks whether the PE is busy or idle. If the scheduled PE is busy, then MET revises the task assignment to alternate PE. Heterogeneous Earliest Finish Time (HEFT)~\cite{topcuoglu1999task} is effective at hierarchical job scheduling. HEFT first sort ready tasks based on the upward rank values, which are importance weights, and greedily map tasks to heterogeneous PEs. An upward rank of a ready task $n_i$ can be recursively calculated by 
\begin{equation}
    rank_u(n_i) = \overline{w_i} + \max_{n_j\in succ(n_i)}(\overline{c_{i,j}} + rank_u(n_j)),
\label{eq:rank}
\end{equation}
where $succ(n_i)$ is a set of successors of task $n_i$, $\overline{c_{i,j}}$ is the average communication cost of edge $(i,j)$, and $\overline{w_i}$ is the average computation cost of task $n_i$. Essentially, the upward rank is the length of the critical path from task $n_i$ to the exit task. While the original research seeks the critical path~\cite{topcuoglu1999task}, a job DAG in DS3 is deemed complete when all of its tasks are finished. Therefore, the performance of HEFT relies heavily on the heuristic task-PE mapping, which iteratively computes the earliest execution finish time (EFT) of a ready task. EFT of task $n_i$ and processor $p_k$ is equated by 
\begin{multline}
    EFT(n_i,p_k) = \max\{ avail[k], \\ \max_{n_j\in pred(n_i)}(AFT(n_j) + c_{i,j})\},
\label{eq:eft}
\end{multline}
where $avail[k]$ is the earliest time at which the processor $p_k$ is ready for task execution, $pred(n_i)$ is the set of predecessor tasks of $n_i$, and $AFT(n_j)$ is the actual finish time of the task $n_j$. $c_{i,j}=w_{i,j}/{B(p_i,p_j)}$, where $w_{i,j}$ is the weight of edge $(i,j)$ and $B$ is bandwidth between given processors, is the data transmission delay as referred to~\eqref{eq:comm_delay}. Essentially, EFT algorithm calculates actual delay-aware computation time and exhaustively schedules the task to the PE with minimal cost. HEFT particularly applies an insertion-based policy that seeks whether the scheduled task can be executed prior to the previous task assignment. If HEFT finds residual gaps due to transmission delays associated with previous scheduling decisions, it reschedules the tasks. Recent improvement via execution-focused heuristic in dynamic run-time scenarios resulted in a run-time variant of HEFT, \texorpdfstring{HEFT\textsubscript{RT}}{heftrt}~\cite{mack2021performant}.

\subsubsection{Neural scheduler}\label{sec:background:scheduler:neural}
DeepSoCS~\cite{sung2020deepsocs} first introduced in the DSSoC with the realistic setting. It sorts tasks using the topological knowledge extracted by graph neural networks and maps each task to PEs using exhaustive search, EFT algorithm~\cite{topcuoglu1999task}, accordingly. DeepSoCS shows a promising result in the SoC application by exploiting the insertion policy and imitating the expert policy, HEFT. However, mapping the tasks to PEs crucially impacts the performance rather than sorting the tasks in DS3 due to counting job completion after all tasks are finished.

SCARL~\cite{cheong2019scarl} is designed for scheduling a single-level job input to heterogeneous machines in a pre-defined number of injecting jobs. SCARL employs attentive embedding~\cite{vaswani2017attention} to share representation from the job and resource embedding and allocate each job to the available machine. The scheduler conducted experiments in the extended version of the simple cluster simulation~\cite{mao2016resource}.

\begin{figure*}[!ht]
\centering
\includegraphics[width=.95\textwidth]{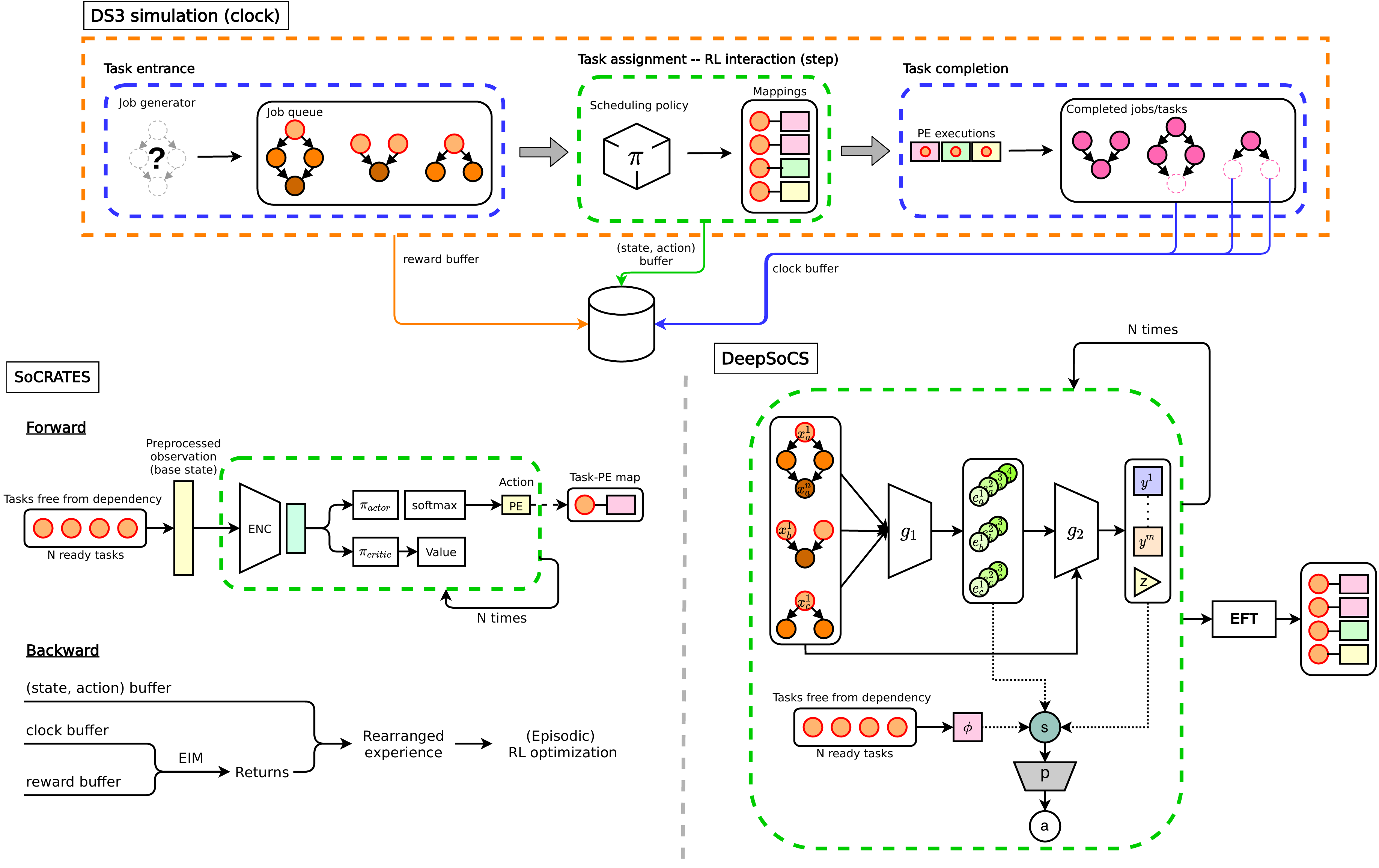}
\caption{\label{fig:overall_struct} The architecture of neural schedulers applied to DS3 simulator. Schedulers receive $N$ tasks in the ready queue and map each task to SoC computing resources. Due to the varying number of tasks, scheduling policies feed each task iteratively. SoCRATES applies Eclectic Interaction Matching to post-process the return (bottom-left). DeepSoCS returns sorted tasks and uses the EFT algorithm to map them to resources (bottom-right).}
\end{figure*}

\section{Proposed method}\label{sec:proposed}
The critical challenges for designing a DRL scheduler in DS3 are that the scheduler requires to i) adaptively allocate a varying number of tasks to heterogeneous PEs by considering system dynamics and data transmission delays, and ii) correctly align task returns to scheduling actions according to respective time-varying agent experiences. An overall systematic workflow of DS3 with scheduling policies is depicted in Fig.~\ref{fig:overall_struct}. After the tasks enter the ready queue, the scheduler receives an observation and maps each task to corresponding PEs. For the following subsections, we provide the state, action, and reward statements tailored to DS3. As the set of actions varies in order and time, we provide cases on how to design actions. Also, we delineate a straightforward and effective EIM technique and how this technique addresses the alignment of return.

\subsection{Agent description} \label{sec:proposed:desc}
Applying RL to sequential decision-making problems is natural, as it collects experiences via interactions with the environment. In general, conventional RL is formalized by Markov Decision Processes (MDP), which is consisted of a 5-tuple $\langle \mathcal{S}, \mathcal{A}, R, P, \gamma \rangle$~\cite{sutton2018reinforcement}. Here, $\mathcal{S} \in \mathbb{R}^d$ is the state space, $\mathcal{A} \in \mathbb{R}^n$ is the action space, and $R \in \mathbb{R}$ is the reward signal that is generally defined over states or state-action pairs. $P: \mathcal{S} \times \mathcal{A} \rightarrow \mathcal{S}$ is a matrix representing transition probabilities to next states given a state and an action. $\gamma \in [0, 1]$ is the discount factor determining how much to care about rewards in maximizing immediate reward myopically or weighing more on future rewards. RL aims to discover an optimal policy $\pi$ that maximizes the expected cumulative (discounted) rewards or (discounted) returns. At every interaction, the RL agent samples a (discrete) action from its policy, which is the probability distribution of actions given a state, $a_t \sim \pi(s_t)$. The agent then computes the return with $\mathbb{E}[\sum^T_{t=0} \gamma^{t-1} R(s_{t}, a_{t})]$, where $t$ is the interaction time step. In this paper, we assume a finite state, finite action, and finite-horizon problem.

\subsubsection{State}\label{sec:proposed:desc:state}
The state representation is designed to capture information of simulation dynamics. Considering the SoC domain-specific knowledge, we select the attributes of the overlapping tasks/jobs and resource information. The observation features at every interaction are 
\begin{multline}
    \text{Concat}((P^G_n, Stat^G_n, TWT^G_n, |pred^G_n|)^{v,W}_{n=0,G=0},\\
    (Dep^G, JWT^G)^W_{G=0}, N_{child}),
\end{multline}
\noindent where $n$ is a task in every job $G$, $v$ is the number of tasks in job $G$, and $W$ is the number of job DAGs in job queue. Each of the observation features is described as follows.
\begin{itemize}
    \item $P^G_n$, the assigned PE ID.
    \item $Stat^G_n$, one-hot embedded task status. Status is classified by one of the labels from ready, running, or outstanding. 
    \item $TWT^G_n$, the relative task waiting time from the ready status to the current time.
    \item $|pred^G_n|$, the number of remaining predecessors.
    \item $Dep^G$, the number of hops (levels) for the remaining tasks as referred to task dependency graph.
    \item $JWT^G$, the relative job waiting time from injected to the system to the execution time.
    \item $N_{child}$, the number of all awaiting child tasks in the outstanding and ready statuses.
\end{itemize}
Time in observation features refers to the actual clock signal in an SoC simulation. Based on the choices of neural architecture designs, state representation includes graph embeddings that capture topological information using graph neural networks~\cite{mao2018learning,cheong2019scarl,sung2020deepsocs}.

\subsubsection{Action}\label{sec:proposed:desc:action}
At every task assignment, shown in the top-middle stage from Fig.~\ref{fig:ds3_workflow}, the agent performs a scheduling decision on an individual task that is free from dependency. Since the number of ready tasks varies by the previous scheduling decisions following their dependencies, the feasible action set varies. Let the ready tasks by $\textbf{a}_t \in \mathcal{T}_{ready}$, where $\mathcal{T}_{ready}$ is a set of ready tasks. For every task $\{a_i\}_{i=1}^{|\mathcal{T}_{ready}|}$, an action $i$ is sampled from the policy distribution with parameter $\theta$, $a_i \sim \pi_\theta(a|s)$, which can be represented by multinomial distribution, $\pi_\theta(a|s) \stackrel{d}{=} \text{Multinomial}(p, m)$. Here, $p \in \mathbb{R}^{1\times Q}$ is the probabilities of each PEs, $Q$ is the number of PEs, and  $m \in \mathbb{R}^{1\times Q}$ is a masking vector to filter out PEs not supporting the task.

One approach is to consider a set of actions as a group action. The group action at RL interaction time step $t$ can be represented by $a_{i,t} \sim \pi_{\theta}(s_t)$, where the set of actions are sampled from the same probabilities with respect to policy distribution. In practice, we define the size of the action vector to a large enough number and apply zero-padding whenever the number of ready tasks is less than that~\cite{vinyals2017starcraft}. An alternative approach is to treat each ready task as an individual action. In lieu of the group action, the agent pulls out each for a PE selection with respect to different policy distribution iteratively, $a_{i,t} \sim \pi_{\theta,(i)}(s_t)$. In this case, each action is sampled from different probabilities with respect to policy distribution.

\subsubsection{Reward}\label{sec:proposed:desc:reward}
DS3 schedulers aim to minimize average latency over simulation length. As described in~\eqref{eq:latency}, the number of completion jobs is largely dependent on the latency. While the negative job duration reward is an adequate reward metric in a cluster environment~\cite{mao2018learning}, this is not effective for latency. Minimizing the elapsed time of the completed jobs is a local optimization, while increasing the number of completed jobs is a global optimization to entail latency minimization in overall. Moreover, maximizing the number of tasks is not adequate optimization because leaving one task out of a job did not contribute to the job completion. We state the reward function as follows.
\begin{equation}
    R(clk) = C_1\cdot |{\hat{\mathcal{G}}_{comp}}| + C_2 \cdot clk,
\label{eq:pos_rew}
\end{equation}
\noindent where $|\hat{\mathcal{G}}_{comp}|$ is the number of newly completed jobs at $clk$, $C_1$ and $C_2$ are the weights of job completion bonus and penalty for clock signal, respectively. Empirically, we design $+50$ for $C_1$ and $-0.5$ for $C_2$. We apply a penalty term to encourage the agent to complete jobs fast. Note that this reward function is computed per clock signal to enable the EIM technique, which is discussed in the following section. For the standard RL approach, the reward is computed per interaction step $t$ instead of clock signal $clk$.

\begin{figure*}[!ht]
\centering
\includegraphics[width=.98\textwidth]{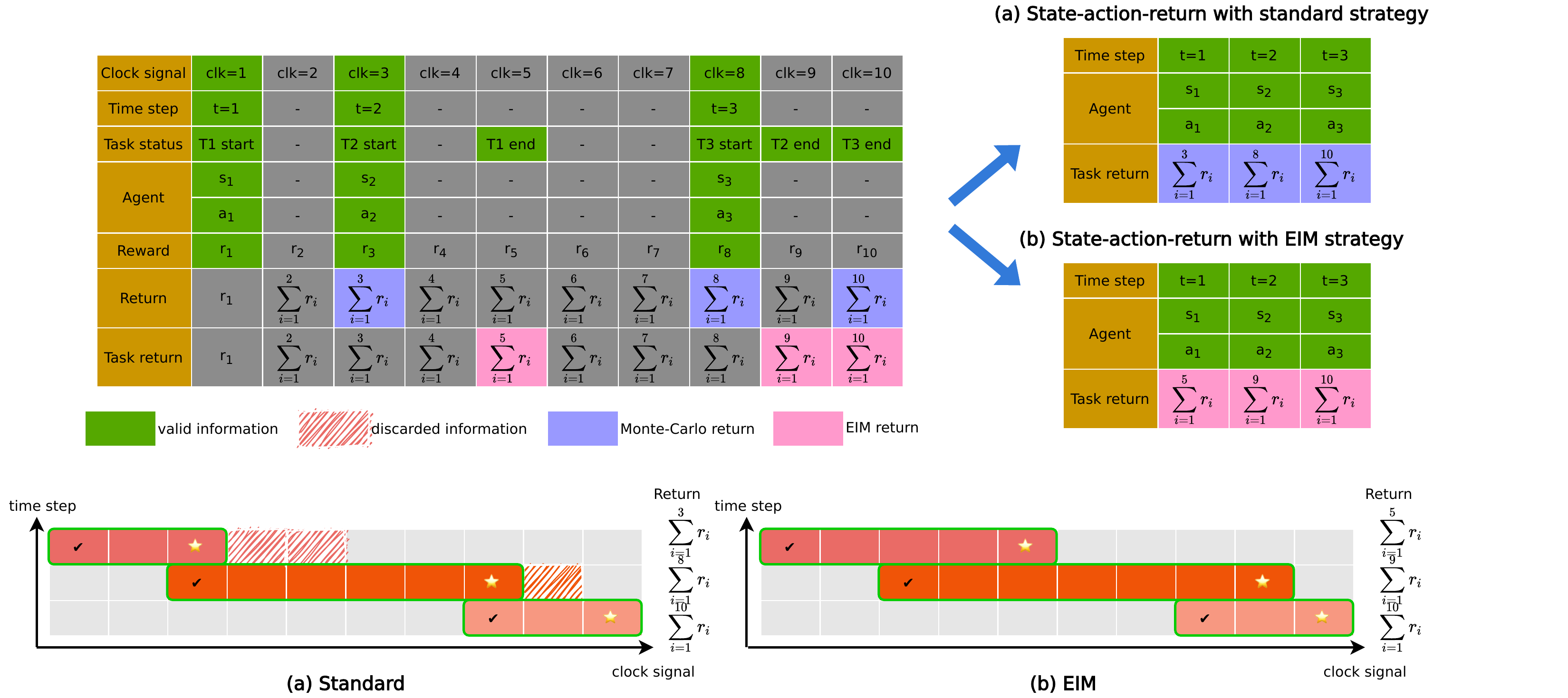}
\caption{\label{fig:eim} An experience with different strategies. Top figure depicts the experiences and rearranged state-action-return sequential triplets after processing different strategies: standard and EIM. EIM preserves the integrity of task execution for return calculation via accounting for returns spanning task duration. Bottom figure describes three tasks coming with concurrency and inconsistent interaction due to task dependency and heterogeneous resources. Two orthogonal axes show interaction time step and actual simulation clock signal. For standard strategy, delayed consequences are discarded, as depicted by shaded regions.}
\end{figure*}

\subsection{Eclectic Interaction Matching} \label{sec:proposed:eim}
Conventional RL environments formalized in standard MDP assumption return the next observation and action consequences right after the previous action has been completed. However, as introduced in the example case in Fig.~\ref{fig:irregular_interaction}, the action in DS3 is performed with the ready tasks, and it is highly not regularly performed due to the variability in task dependency from a mix of incoming jobs. As a result, the next observation is not immediately generated after executing the previous scheduled task. Moreover, treating a reward and the next observation at the same time leads to incorrect reward propagation, because the scheduler assigns multiple tasks at the same time, and each of the scheduled tasks readily be completed at different time due to the different performance of heterogeneous PEs. In that sense, the task dependencies and different task duration inherently cause delayed rewards, and this phenomenon leads to incorrect reward propagation in the optimization updates. Therefore, the scheduling agent must handle a varying number of action sets and the mismatches between the interaction and the action effects during the action decision stage, for which we address right below.

A standard RL experience comes down to a sequence of $\langle s, a, r, s'\rangle$. We first decouple the receiving reward and next state to have i) a sequence of $\{ s_i, a_i \}^{T}_{i=1}$, where $T$ is the last interaction step, and ii) a list of rewards collected upon the simulation clock signal $\{r_{clk}\}_{clk=1}^{CLK} = \{R(clk)\}^{CLK}_{clk=1}$, where $R(clk)$ is a reward function described in~\eqref{eq:pos_rew}. As discussed in Section~\ref{sec:background:analysis}, an amount of interactions $T$ and entire clock signal $CLK$ are not generally matched due to the different task dependencies in a mix of hierarchical jobs and performances in heterogeneous PEs. We compute the immediate reward per clock signal independent of the interaction step. Additionally, we append the `start' flag to the state-action tuple and stored in the buffer. While traversing the simulation, at the completion of any scheduled task, we store the `complete' flag and completed clock signal $\omega$ in the buffer. Hence, the experiences in the buffer is described as $\{ s_t, \{a_{(n,t)}\}_{n=1}^{n'}, \{\omega_{(n,t)}\}_{n=1}^{n'} \}_{t=1}^{T}$, where $n'$ is the number of ready tasks at interaction step $t$.

Fig.~\ref{fig:eim} showcases an exemplary experience of scheduling three ready tasks and return computations using two different strategies. On the upper-left diagram, an agent receives an observation and sequentially selects an action. The state, action, and task starting/completing clock signal are marked green. Next, we compute returns based on the accumulated rewards. In the standard approach performing the Monte-Carlo return with the accumulated rewards~\cite{sutton1999policy}, partial reward sequences that overlap ongoing tasks and subsequent observation are not counted; the missing sequences incur incorrect return matching and instability in training.

The EIM technique instead aligns Monte-Carlo returns with the committed actions spanning individual task duration, referred to the `start' and `end' task signals. The return for each action reflects the length of task duration, and each action correctly matches outcomes without any discarded information. Moreover, task flags and actual clock signals allow the agent to sequentially select actions within a set of varying actions. EIM technique thus enables the agent to receive a correct form of state-action-return triplets, regardless of varying action sets. EIM is a straightforward post-process that is proven effective in training an agent when the agent interaction and simulation clock signal is inconsistent. The bottom diagram of Fig.~\ref{fig:eim} depicts the task and action with the return computation. The x-axis denotes the simulation clock signal, and the y-axis is the RL interaction time step. Partitions of second and third task duration in the standard approach are discarded for return assignment. EIM, by contrast, properly pairs the state-action tuple with returns by aligning returns to task assignments.

\begin{algorithm}[!t]
\caption{SoCRATES Scheduler}
\begin{algorithmic}[1]
\STATE {\bfseries Input:} clock signal $clk$, job queue $\mathcal{J}$, ready task queue $\mathcal{T}_{ready}$
\FOR{each episode}
\STATE state-action buffer $\mathcal{B}_{SA} \leftarrow \emptyset$
\STATE clock buffer $\mathcal{B}_{clk} \leftarrow \emptyset$
\STATE reward buffer $\mathcal{B}_{R} \leftarrow \emptyset$
\FOR{each task $i$ in $\mathcal{T}_{ready}$}
\STATE Construct state $s_{t}$
\STATE $a_{i,t} \sim \pi_{\theta,(i)}(s_{t})$
\STATE Assign $a_{i,t}$ to PE for task $i$
\STATE $\mathcal{B}_{SA} \leftarrow (s_{t}, a_{i,t})$
\ENDFOR
\IF{task $i$ complete}
\STATE $\mathcal{B}_{clk} \leftarrow (i,\omega)$
\ENDIF
\STATE $\mathcal{B}_R \leftarrow r_{clk}$
\ENDFOR
\STATE \textcolor[rgb]{0.5,0.5,0.5}{\# Update the agent model}
\STATE $\theta \leftarrow \theta + \eta \gamma^t \nabla_\theta \mathcal{L}^{SoC}_t(\theta)$
\item[]
\end{algorithmic}
\label{alg:socrates_pseudo}
\end{algorithm}

In training, we use the Actor-Critic algorithm~\cite{konda1999actor}. We use shared neural networks on both actor and critic and update parameters with REINFORCE~\cite{sutton1999policy}. While the actor network selects actions with respect to the policy distribution, $\pi_\theta$, the critic network estimates the value using the value function, $\hat{V}^\pi_\theta$. At the end of the episode, EIM post-processes the expected returns as described in~\eqref{eq:eim-exp-return}: 
\begin{equation}
    \hat{G}(s_t, \omega) = \sum^{\omega}_{clk=0}\gamma^{clk} r_{clk+1}.
\label{eq:eim-exp-return}
\end{equation}
The actor loss is equated by
\begin{equation}
    \mathcal{L}^{ACT}_t(\theta) = -\sum^T_{t=0}\log \pi_\theta(a_t|s_t)\Big[ \hat{G}(s_t, \omega) - \hat{V}^\pi_\theta(s_t) \Big],
\end{equation}
\noindent and the critic loss is computed by the standard mean squared loss,
\begin{equation}
    \mathcal{L}^{CRI}_t(\theta) = \frac{1}{2}\Big(\hat{G}(s_t, \omega) - \hat{V}^\pi_\theta(s_t)\Big)^2.
\end{equation}
The overall loss is given as:
\begin{equation}
    \mathcal{L}^{SoC}_t(\theta) = \mathcal{L}^{ACT}_t(\theta) + \mathcal{L}^{CRI}_t(\theta) + \xi\mathcal{H}(s_t),
\end{equation}
\noindent where the last term on the right-hand side is the entropy regularization, $\mathcal{H}(s_t) = \mathbb{E}_{\pi_\theta}[\log\pi_\theta(s_t)]$, with its coefficient $\xi$ introduced for exploration. Pseudocode for the proposed algorithm is given in Algorithm~\ref{alg:socrates_pseudo}.

\section{Evaluation}\label{sec:eval}

This section demonstrates the feasibility of neural schedulers in a high-fidelity SoC simulation, DS3. We present evaluations in three ways: (a) we revisit rule-based schedulers and observe their benefits on performance, (b) we verify the efficacy of EIM technique on neural schedulers by investigating PE usage with various reward functions and different action designs, and (c) we empirically validate that neural schedulers can have competitive and generalized performance on run-time overhead in a series of experiments where job DAG topology and PE performance are varied. Specifically, we examine in which operational conditions existing neural schedulers with EIM have significant performance gains.

\subsection{Experimental setup} \label{sec:eval:setup}

\begin{table}[!t]
\centering
\begin{tabular}{ cc }
Hyperparameter & Value \\
\hline
Optimizer & Adam \\
$\gamma$ (discount factor) & 0.98 \\
$\eta$ (learning rate) & 0.0003 \\
$\xi$ (entropy coefficient) & 0.01 \\
$\alpha$ (job structure) & 0.8 \\
$\mu$ (scale to PE performance) & 0.5 \\
$\nu$ (avg. comm. delay) & 0.0 \\
Number of workloads & 200 \\
Simulation length & 10,000 \\
$C$ (job capacity) & 3 \\
Gradient clip & 1 \\
Scale & 25 \\
\hline
\end{tabular}
\caption{\label{tbl:train_const} A table of hyperparameters used for training neural schedulers}
\end{table}

Table~\ref{tbl:train_const} describes a list of training parameters. We use Adam optimizer~\cite{kingma2014adam} and clip the gradients to avoid gradient explosion. To engender more interactions and a more dynamic environment, we randomly inject jobs with a set of 200 workloads. As we empirically discovered that the job DAG topologies of synthetic or real-world profiles are structured with $\alpha = 0.8$, we synthesize job structures with $\alpha=0.8$ based on the given job profile. The job inter-arrival rate (scale) is set to 25; the system stochastically injects a job at every 25 clock signals on average. At every episode, the simulation executes until 10,000 clock signals. To reduce the training and evaluation time, we conduct all experiments with the initial condition of quasi steady state, that is, each experiment begins with all jobs already stacked in the job queue~\cite{sung2020deepsocs}. All evaluations have been conducted by 20 trials with different random seeds.

\begin{figure*}[!t]
\centering
\includegraphics[width=0.95\textwidth]{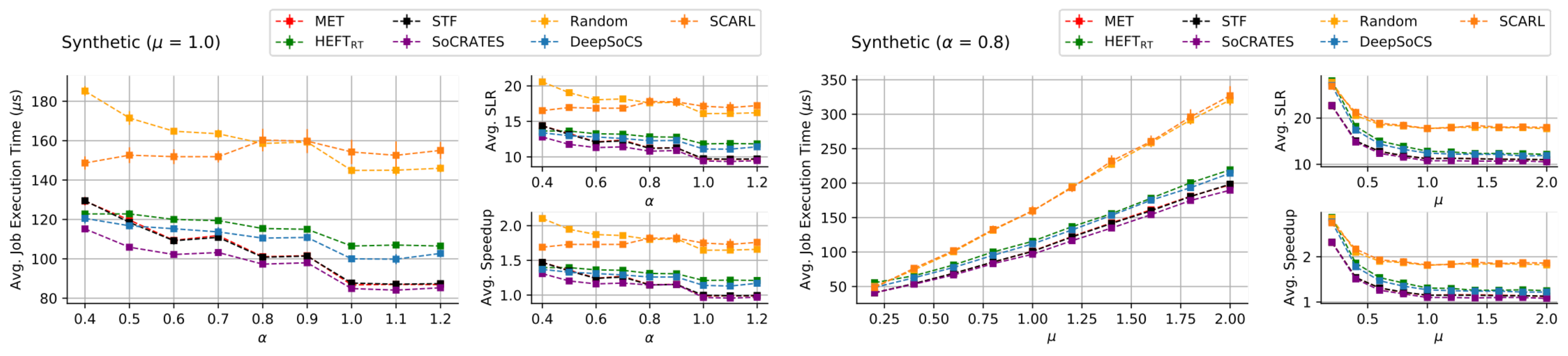}
\caption{\label{fig:syn_results} Performances evaluation on neural and non-neural schedulers with various job structures and PE performances using synthetic profiles. The left and right figures show run-time performance on varying topologies in job DAGs and PE performances, respectively.}
\end{figure*}

\subsection{Revisiting rule-based schedulers}\label{sec:eval:rule-based}
Rule-based algorithms have continue to demonstrate state-of-the-art performance in SoC run-time scheduling~\cite{arda2020ds3,mack2021performant}. In order to establish a baseline for comparative study, first we extensively investigate the run-time performance of existing heuristic schedulers. As described in Section~\ref{sec:background:scheduler:rule}, the main discrepancy between STF and MET is that MET reschedules the scheduling assignment by checking whether the selected PE is busy or idle. Likewise, \texorpdfstring{HEFT\textsubscript{RT}}{heftrt} iteratively computes the actual run-time of given tasks using computation time and data transmission delay. It then applies an insertion policy, which exhaustively searches for a possible empty slot between each task assignment. 

\begin{figure}[t!]
\centering
\includegraphics[width=0.36\textwidth]{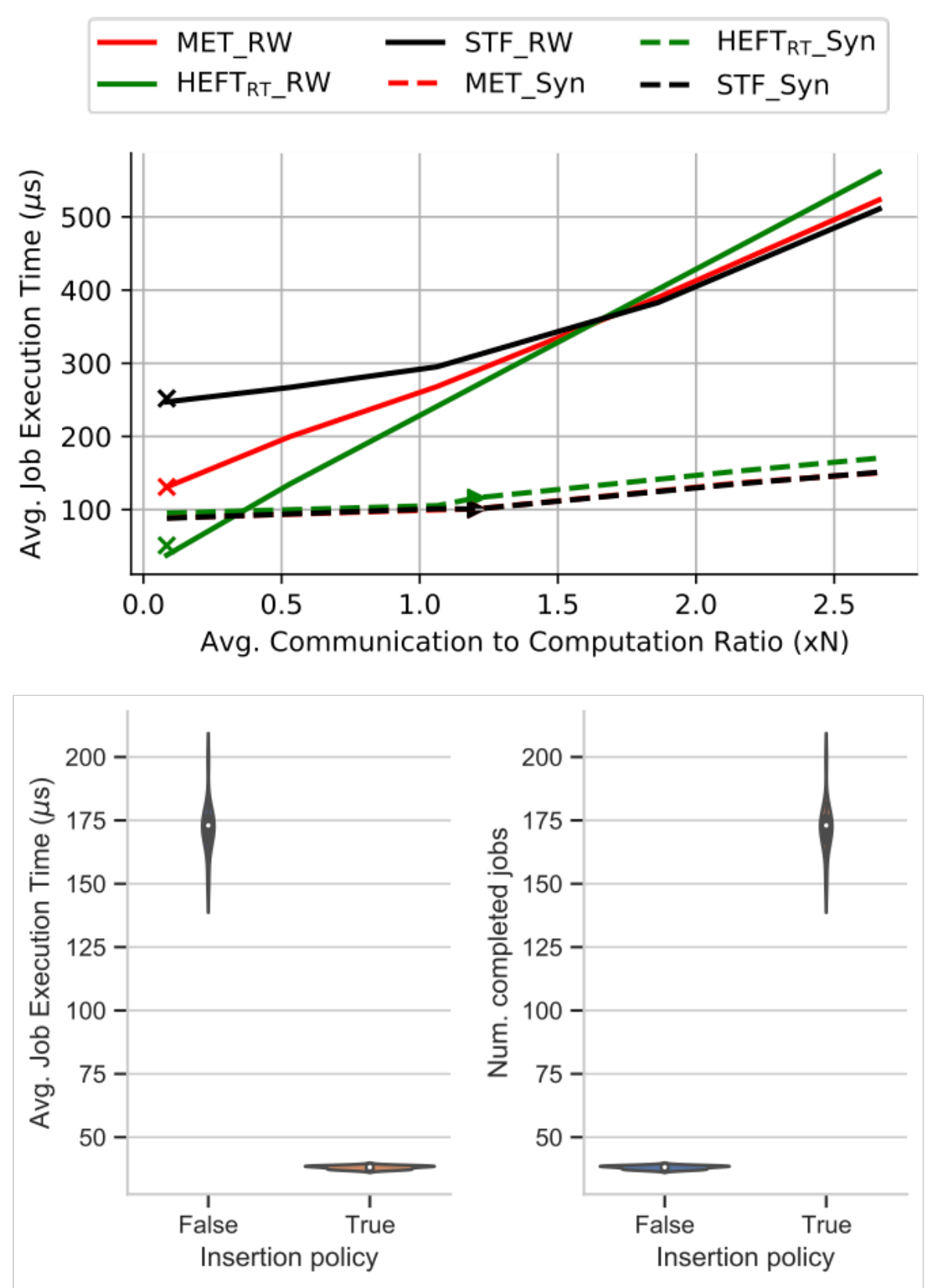}
\caption{\label{fig:heuristics_results} An experimental analysis of heuristic schedulers. The top figure compares run-time performance with different data transmission delays. The cross and triangle marks depict fixed job profiles from synthetic and real-world workloads. We report the results with a solid line for real-world (RW) and a dotted line for synthetic (Syn) workloads for the increase in data transmission delay compared to task computation time. Both lines are plotted by varying the job structure with $\alpha=0.8$. The bottom figure shows the effectiveness of insertion policy in \texorpdfstring{HEFT\textsubscript{RT}}{heftrt} scheduler using violin chart. An insertion policy significantly improves performance for average latency (bottom-left) and increases the number of completed jobs (bottom-right).}
\end{figure}

The top plot in Fig.~\ref{fig:heuristics_results} shows an overall run-time performance of different heuristic schedulers using synthetic (Syn) and real-world (RW) profiles. The x-axis indicates $CCR$, and the y-axis indicates the average latency. The jobs are communicative-intensive, if $CCR \gg 1.0$, and are computation-intensive, if $CCR \ll 1.0$. For synthetic profiles structured with a similar range of $CCR$, STF and MET have comparable performance and surpass \texorpdfstring{HEFT\textsubscript{RT}}{heftrt}. On the other hand, MET and \texorpdfstring{HEFT\textsubscript{RT}}{heftrt} significantly improve performance for real-world profiles, where jobs are computation-intensive, by checking the availabilities in PEs and exhaustively searching the empty slot between task assignments. Particularly, when increasing $CCR$ on real-world profiles, MET and STF show similar performance and surpass \texorpdfstring{HEFT\textsubscript{RT}}{heftrt}. We observe that the increasing gap between computation time and communication delay leads to large variances in the distribution of task run-time. The high variances in profile statistics result in more chances to improve the performance by rescheduling task assignments.

The bottom plot in Fig.~\ref{fig:heuristics_results} shows an experimental result for real-world profile using two types of \texorpdfstring{HEFT\textsubscript{RT}}{heftrt}, with and without insertion policy. The insertion policy effectively seeks better placements due to the divergent distribution of task computation time. Hence, the rescheduling task assignment in the heuristic schedulers instrumentally improves run-time performance. In that sense, rescheduling task assignments can largely improve performance, and \texorpdfstring{HEFT\textsubscript{RT}}{heftrt} can show almost optimal performance within a myopic scope by its exhaustive search when the variations in task run-time are large.

\begin{figure*}[!t]
\centering
\includegraphics[width=0.95\textwidth]{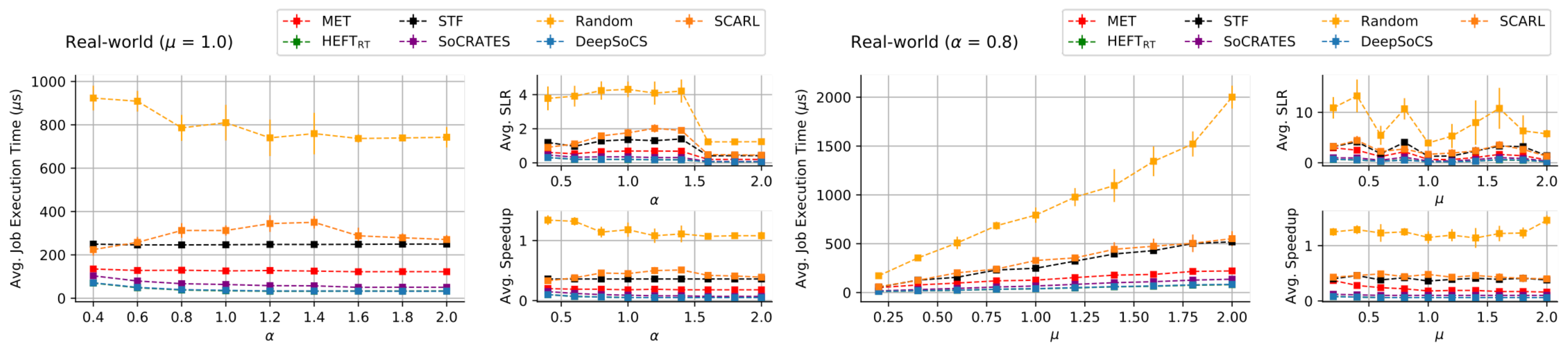}
\caption{\label{fig:wifi_results} Performances evaluation on neural and non-neural schedulers with various job structures and PE performances using real-world profiles. The left and right figures show run-time performance on varying topologies in job DAGs and PE performances, respectively.}
\end{figure*}

\subsection{Performance comparison} \label{sec:eval:performance}
This section describes our extensive evaluation of the performance of existing schedulers specifically designed for heterogeneous resources in the SoC domain. We compare two types of representative scheduling algorithms: i) SoCRATES~\cite{sung2021socrates}, DeepSoCS~\cite{sung2020deepsocs}, and SCARL~\cite{cheong2019scarl} for neural, and ii) STF, MET~\cite{braun2001comparison}, and \texorpdfstring{HEFT\textsubscript{RT}}{heftrt}~\cite{topcuoglu1999task,mack2021performant} for heuristics. Since SCARL does not support hierarchical workloads, we modified SCARL as follows: (1) \textit{State}: We take the same job representation with the SoCRATES. We select PE performance, types of PE, capacity, available time to execute tasks, task remaining execution time, idle rate, and normalized values of PE run-time and expected total task time for features of PE representation. (2) \textit{Action}: Original SCARL selects both workload and resource. Since SCARL does not support task selections for hierarchical workloads, the action maps the selected resource to the task in sequence. (3) \textit{Reward}: We use the same reward function of job completion, described as~\eqref{eq:pos_rew}. At the update stage, we compute the returns with the collected rewards after post-processing with EIM.

Throughout the evaluations, we primarily concentrate on \textit{average latency}, which indicates the average run-time performance. We observe how the schedulers behave in a wide range of experiments by varying the types and structures of the jobs, transmission delay, and performance in heterogeneous PEs. Fig.~\ref{fig:syn_results} reports the run-time performance using a synthetic workload. The right and left plots depict the experimental results after varying job structures and PE performance. For the former case, we control the job structure parameter $\alpha$ while holding the parameter of PE performance $\mu$, and for the latter case, vice versa. Large $\alpha$ generates shallow but wide job graphs, while small $\alpha$ generates deep but narrow job graphs. All evaluations are conducted with the highest job inter-arrival rate (the smallest scale value), leading to a high frequency of job injection. From the holistic viewpoint, the trends in SLR and Speedup follow the curve of the average latency. Among all other schedulers, we can observe that SoCRATES has surpassed under a wide range of experiment settings. Since the neural schedulers have been evaluated using a single trained model, SoCRATES has generalized and competitive performance in various scenarios in job structures and PE performances. As described in Section~\ref{sec:eval:rule-based}, $CCR$ for the synthetic workload closely reaches 1.0, meaning that the task computation time and data transmission delay lie in a similar range. As a result, the task ordering in heuristics did not impact much, and their performances fell behind the SoCRATES. When the number of tasks was varied, SCARL's attentive embedding of tasks and resources was unable to take advantages of attentive representation and even further deteriorates the overall run-time performance. As a result, SCARL shows comparable performance to random policy. 

Synthetic and real-world profiles differ in the number of tasks and resources, job DAG topology, and supported functionalities on individual resources. Table~\ref{tbl:diff_envs} indicates that the real-world profile has a much higher task computation time cost than the data transmission delay. Hence, the actual task run-time is more varied, and rescheduling task assignments from heuristic schedulers can largely improve the run-time performance. As a result, SCARL significantly improves performance, and \texorpdfstring{HEFT\textsubscript{RT}}{heftrt} shows the most optimal run-time behavior in the real-world profile, as shown in Fig.~\ref{fig:wifi_results}. We observe that SoCRATES surpassed other schedulers in the synthetic profile, but it was limited in the real-world profile. Although EIM remedies fundamental difficulties in DS3 and improves SoCRATES performance, it cannot reduce the performance gap for the optimal task scheduling in a myopic range using an exhaustive search. We hypothesize that the characteristics of the real-world profile, such as various availabilities of task executions in resources, invokes cohesive challenges to designing a DRL scheduler. Additionally, a large number of tasks leads to increased complexity in task dependency composition and large variance because completing all tasks counts as job completion. Hence, the end-to-end neural approach could not surpass \texorpdfstring{HEFT\textsubscript{RT}}{heftrt} when the task duration has high variance and computing resources cannot compute all tasks. Although SoCRATES shows limited performance in the real-world profile, DeepSoCS shows comparable performance to \texorpdfstring{HEFT\textsubscript{RT}}{heftrt} by imitating experiences from the expert algorithm.

\begin{figure}[!t]
\centering
\includegraphics[width=0.48\textwidth]{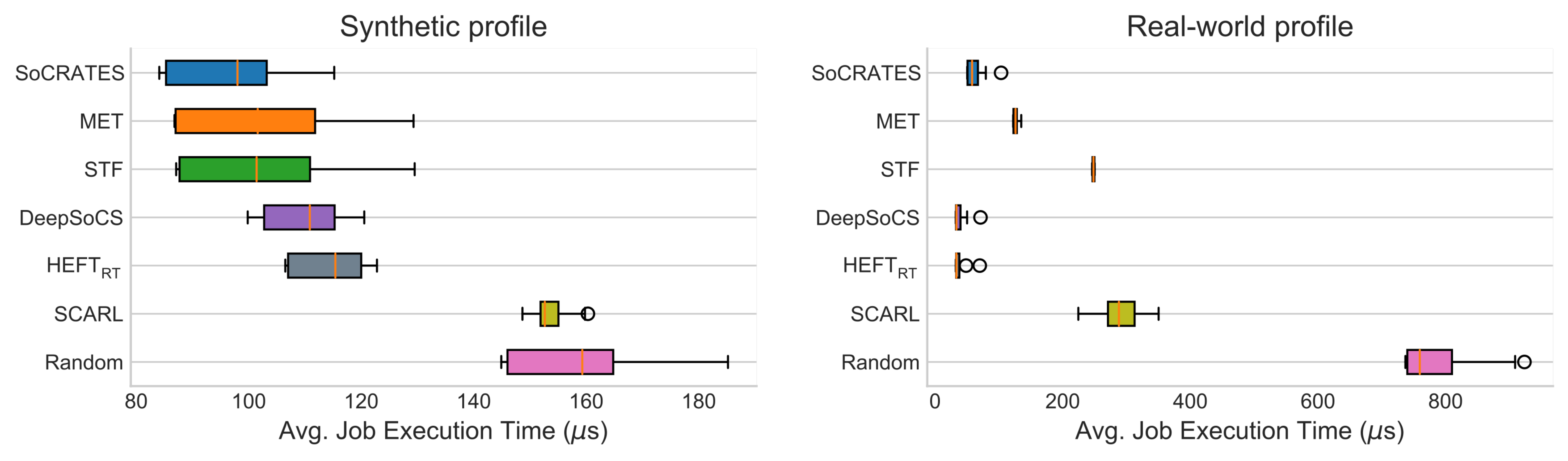}
\caption{\label{fig:fin_eval} An overall result of \textit{average latency} with various job DAG topology by controlling $\alpha$. The left plot shows the synthetic profile, and the right shows the real-world profile.}
\end{figure}

In conclusion, Fig.~\ref{fig:fin_eval} demonstrates the overall evaluation of neural and non-neural schedulers using synthetic and real-world profiles with various job DAG topology by controlling $\alpha$. The left figure shows that SoCRATES has largely improved behavior when the computation time and communication delay lie in a similar range. The right figure shows that DeepSoCS and \texorpdfstring{HEFT\textsubscript{RT}}{heftrt} shows the most optimistic performance when the composition of task duration has a large variance. Thus, if we adaptively choose a neural scheduler between EIM-based policy and imitated expert policy depending on different scenarios, the neural SoC schedulers can obtain an improved performance over other neural and non-neural schedulers.

\subsection{Anatomy of SoCRATES} \label{sec:eval:anatomy}
SoCRATES is the fully differential decision-making algorithm~\cite{sung2021socrates}. The crucial component in SoCRATES is EIM technique that alleviates both delayed rewards and variable action selection, caused by hierarchical job graphs, mixes of different jobs, and heterogeneous computing resources. Although the recently introduced EIM technique overcoming such challenges, it lacks the validation of the efficacy of EIM. In this section, we rationalize the underlying reasons behind the performance improvement of EIM by examining PE usages and action designs.

\subsubsection{Analysis of Eclectic Interaction Matching} \label{sec:eval:anatomy:eim}
\begin{figure}[t!]
\centering
\includegraphics[width=0.41\textwidth]{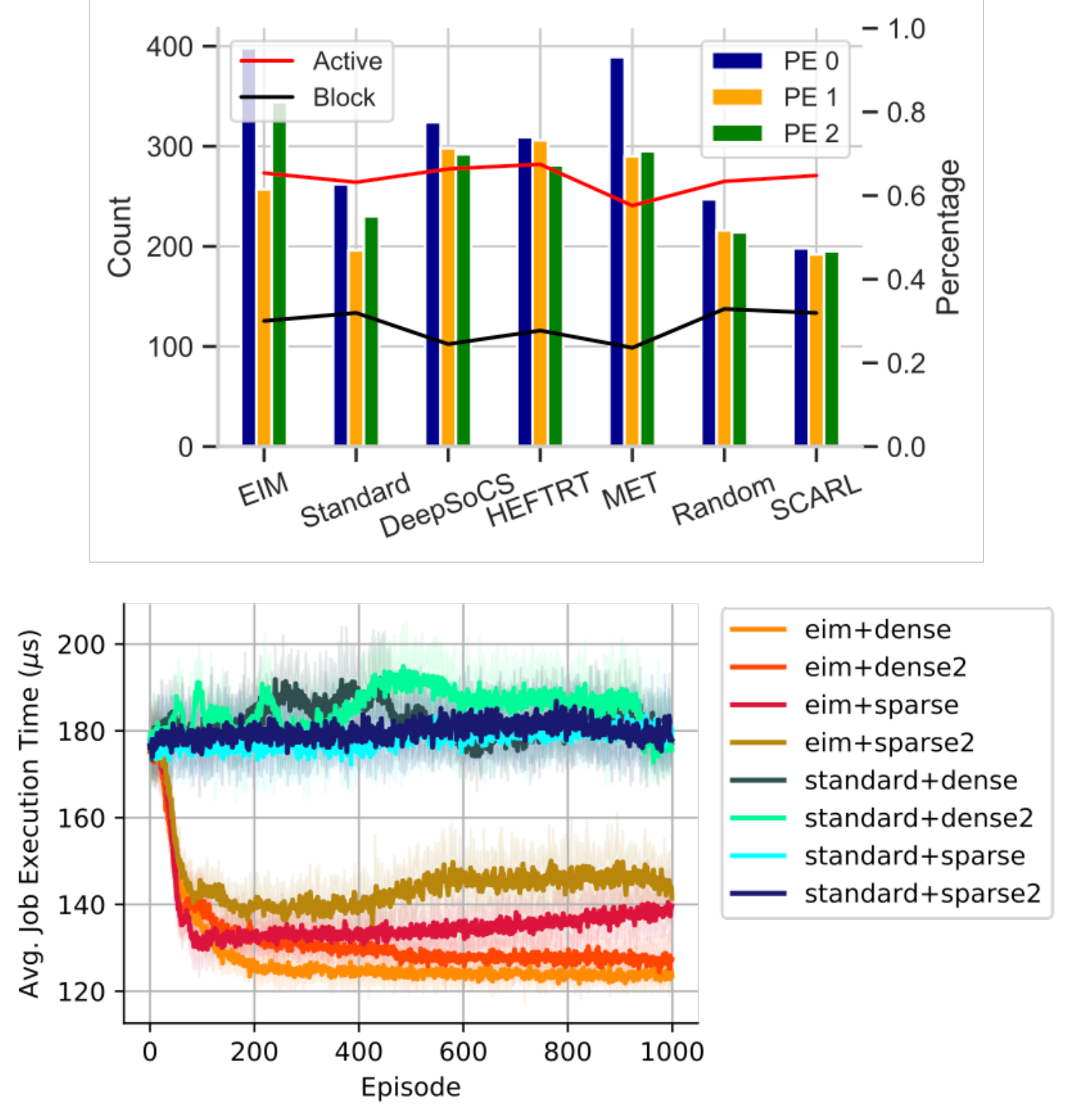}
\caption{\label{fig:eim_analysis_results} Experiments for the analysis of EIM technique. Top plot shows PE selection and active/block time for PEs in different scheduling algorithms. Bottom plot compares latency performance using EIM and standard strategies with various sparse and dense reward functions. All experiments are conducted with synthetic workloads with a fixed job topology.}
\vspace{-6.3mm}
\end{figure}

First, we examine how EIM affects the scheduling policy decisions with the PE selection behavior. The top plot in Fig.~\ref{fig:eim_analysis_results} shows the counts on each PE execution and the total amount of time for PE in active and blocking time. Active time represents the amount of time in PE execution, and blocking time represents the duration when a PE is busy while the assigned tasks are ready. The simulation clock signal measures each time. Intuitively, optimized performance in PE usage can be achieved when active time increases but blocking time decreases. SoCRATES with EIM technique, in effect, utilizes a greater number of PEs and achieves higher resource active time than that with other policies. Its high blocking time derives from the fact that the policy weights encourage achieving long-term returns while blocking the cost of immediate returns. The same policy without EIM technique also has similar values of active and blocking times. However, its low number of PE counts leads to poor PE utilization and latency behavior. MET also uses a large quantity of PEs, but its low active time in PEs invokes additional bottlenecks in PE usage. Random and SCARL policies show high value in active time. However, their absolute number of PE counts is much lower. As a result, they have poor performance compared to other schedulers. 

Next, as shown in the bottom plot, we train SoCRATES using various types of reward functions with and without the EIM technique for generalization. We train the policy using synthetic workload, and each two types of dense and sparse reward functions are used (see Appendix~\ref{sec:append:reward_fn_eim} for more details). The solid line represents the average values, and the shaded region bounds the maximum and minimum values among 8 runs in random seeds. The standard strategy seemingly cannot train the model by observing its steady and straight performance curve. On the other hand, the EIM strategy enables to show learning curve. The EIM iteratively matches each return in actions with respective task duration at the cost of storing extra flags on task start and completion. This additional post-processing is very cheap in operations and achieves substantially better latency performance in any kind of reward function than the standard strategy. From the reward function perspective, the sparse reward apparently exacerbates unstable latency performance due to its limited feedback for an RL agent. Hence, it is commonly modified to dense forms using the shaping technique~\cite{ng1999policy}. 

\begin{figure}[t!]
\centering
\includegraphics[width=0.46\textwidth]{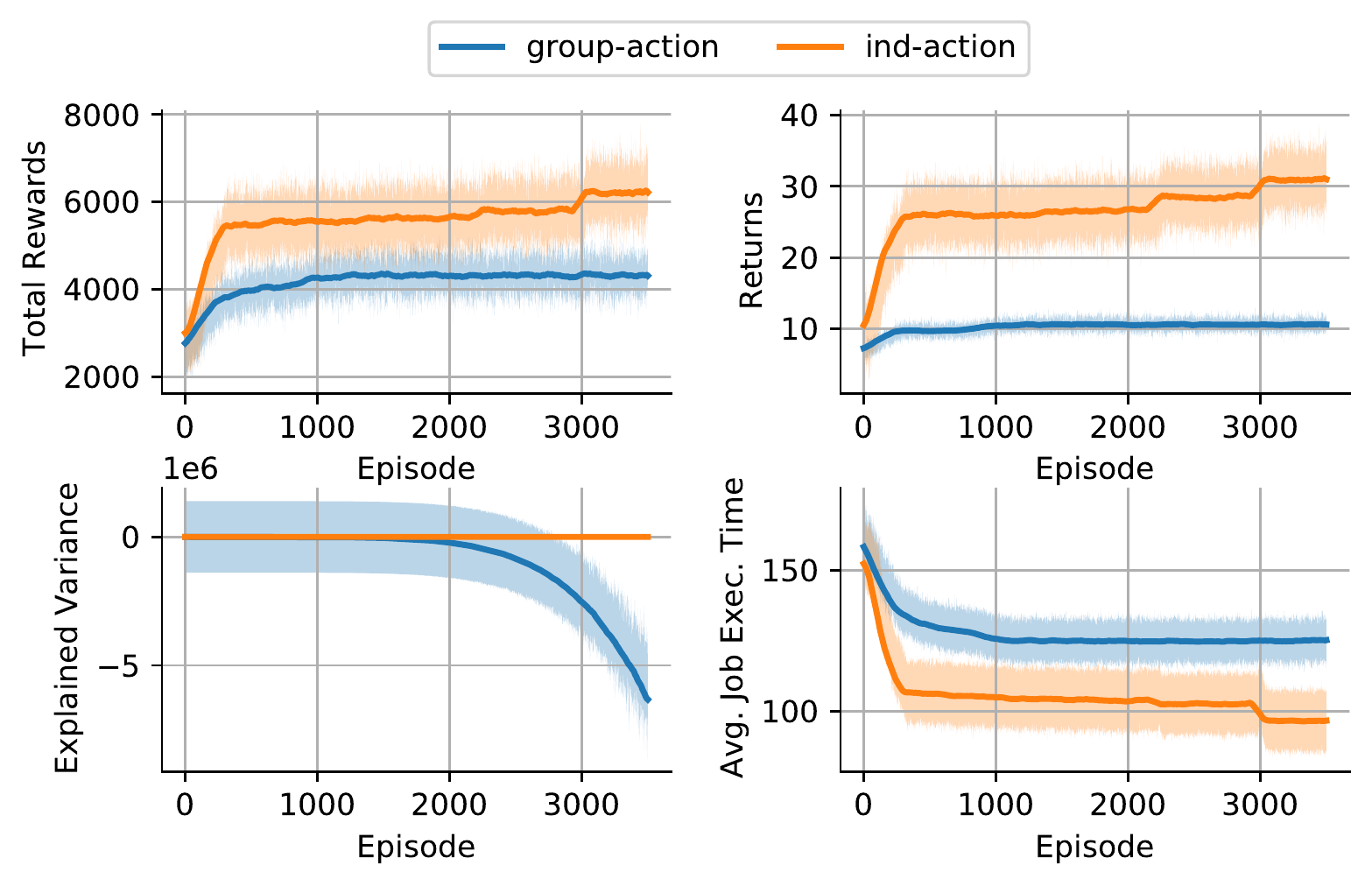}
\caption{\label{fig:group-vs-indep} A set of performance metrics for selecting a group action and independent actions in the training phase. Top-left: total rewards, top-right: returns, bottom-left: explained variance~\cite{dhariwal2017openai}, and bottom-right: average job execution time ($\mu s$)}
\vspace{-6.3mm}
\end{figure}

\subsubsection{Action design} \label{sec:eval:anatomy:action}
To design a DRL agent in an environment with varying actions, one can set an action space to the number of maximal actions and mask out every varying action~\cite{vinyals2017starcraft}. In the case of group actions, we distribute the returns, computed by the longest task duration, to the set of actions. It turns out, however, that the approach to group actions is inadequate due to the rapid convergence of gradients. The action probabilities quickly devolve to the local minima, and the losses in policy and value increase exponentially. Also, a set of group actions and their respective returns cannot be distributed to individual actions, since each task has a respective reward and estimated return or value in the task duration. On the other hand, individual action selection with the EIM technique converts a varying action problem into a conventional sequential decision-making problem, with no need to be aware of invalid actions from the agent perspective. Fig.~\ref{fig:group-vs-indep} shows that the policy with independent actions produces much higher values in the expected returns by 270\% and has improved run-time performance by 30\%. We applied the EIM technique to both approaches. Additionally, we report the explained variance, $EV$, denoted by~\eqref{eq:ev}. 
\begin{equation}
    EV(s_t) = \frac{1-\mathbb{V}[G(s_t) - \hat{G}(s_t)]}{\mathbb{V}[G(s_t)]},
\label{eq:ev}
\end{equation}
\noindent where $G(s_t)$ is empirical return of state $s_t$ and $\hat{G}(s_t)$ is predicted return of state $s_t$. $EV$ measures the difference between the expected return and the predicted return~\cite{baselines}. By observing the decreases in explained variance for the group action, we empirically validate that group action does not fully understand the environment while reconciling the experiences.

\section{Conclusion and future work}\label{sec:concl}

In this paper, we unveil myths and realities of  DRL-based SoC job/task schedulers. We identify key practical challenges in designing high-fidelity neural SoC schedulers: (1) varying action sets, (2) high degree of heterogeneity in incoming jobs and available SoC compute resources, and (3) misalignment between agent interactions and delayed rewards. We propose and analyze a novel end-to-end neural scheduler (SoCRATES) by detailing its core technique (EIM) which aligns returns with proper time-varying agent experiences. EIM successfully addresses the aforementioned challenges, endowing SoCRATES with a significant gain in average latency and generalized performance over a wide range of job structures and PE performances. We also rationalize the underlying reasons behind the substantial performance improvement in existing neural schedulers with EIM by examining actual PE usages and disparate action designs. Through extensive experiments, we discover that $CCR$ significantly impacts the performance of neural SoC schedulers even with EIM. At the same time, we find that the action of rescheduling task assignments by heuristic schedulers leads to significant performance gain under certain operational conditions, often outperforming neural counterparts. 

With these findings, we intend to investigate further whether EIM technique can bring additional performance gains in other learning-based and planning algorithms, both empirically and theoretically. With the advantage of task rescheduling in heuristic schedulers, we plan to improve neural schedulers by converting such technique to a differential function and integrating it into the optimization. Alternatively, offline reinforcement learning using expert or trace replay~\cite{agarwal2020optimistic} is another possible approach to improve neural schedulers. Moreover, leveraging the structure of the underlying action space to parameterize the policy is a candidate approach to tackle a varying action set~\cite{chandak2020lifelong}. We also plan to leverage GNNs to bestow the structural knowledge from job DAGs~\cite{yu2019dag}, and demonstrate the performance gain of the improved neural schedulers by using the Compiler Integrated Extensible DSSoC Runtime (CEDR) tool, a successor to DS3 emulator, as it enables the gathering of low-level and fine-grain timing and performance counter characteristics~\cite{mack2022cedr}.

\appendices
\section{Job DAG construction} \label{sec:append:job_construct}
\begin{figure}[!t] 
\centering
\includegraphics[width=0.45\textwidth]{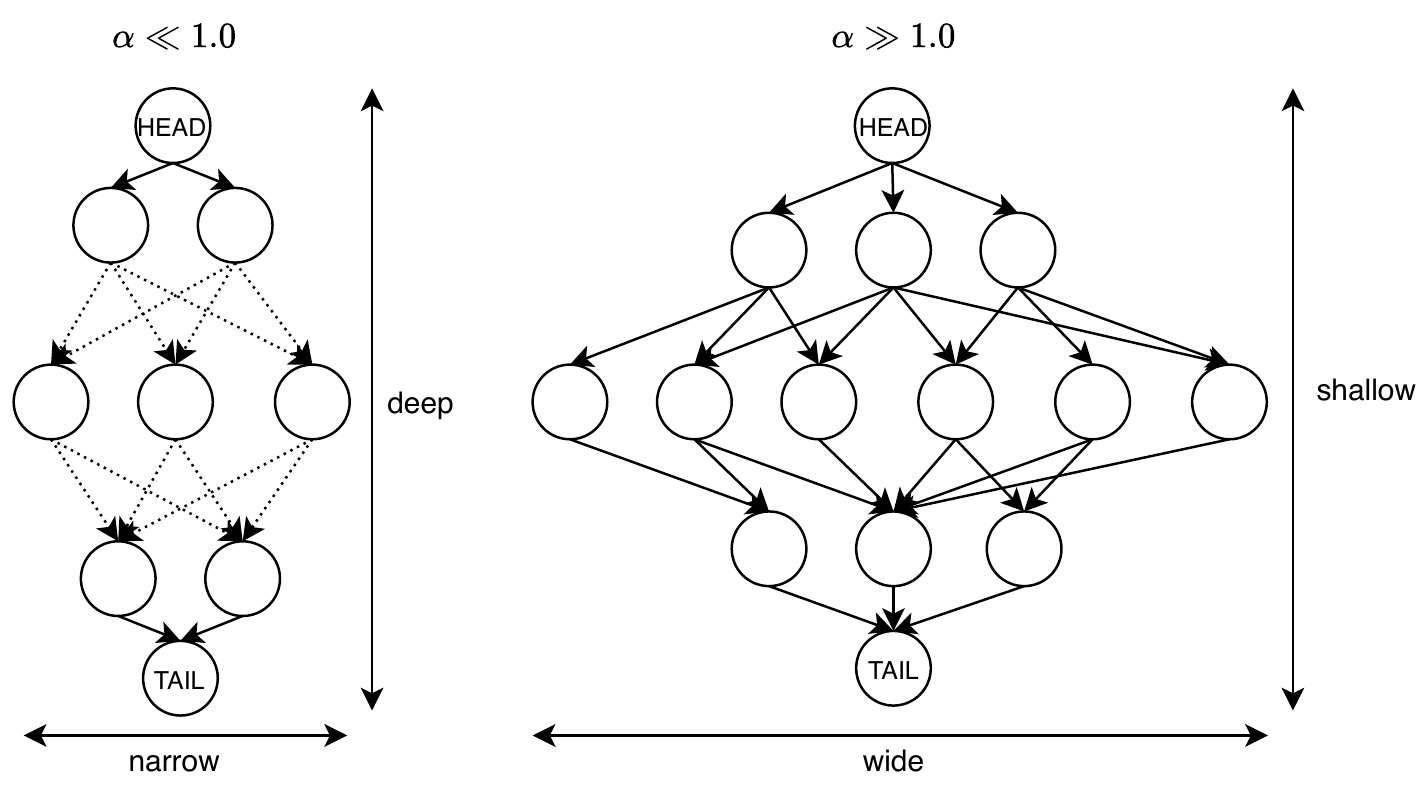}
\caption{\label{fig:alpha_desc} An illustration of two types of job DAGs based on $\alpha$. Left diagram shows that small $\alpha$ generates a deep but narrow job graph. Dotted lines in the middle represent hidden nodes and edges. Right diagram shows that large $\alpha$ generates a shallow but wide job graph.}
\end{figure}

The simulator can synthesize a variety of workloads given the job profile and hyperparameters, which are described in Section~\ref{sec:background:sim:job}. First, we compute average values of widths, $\overline{w}$, and depths, $\overline{d}$, with the hyperparameter $\alpha$ based on the job model description in Section~\ref{sec:background:sim:job}. We compute the number of nodes by $w\sim \max(1, \mathcal{N}(\lfloor \overline{w} \rfloor, 0))$ per $\overline{d}-2$ job levels. Here, we exclude two levels in which the HEAD and TAIL nodes are located. Then we check whether the total number of nodes matches $v$ (the number of nodes). If the total number of nodes is less or greater than $v$, then we randomly select nodes from the job DAG and add/subtract them in order to exactly have $v$ nodes. As illustrated in Fig.~\ref{fig:alpha_desc}, small $\alpha$ generates deep but narrow job graphs (left figure), and large $\alpha$ generates shallow but wide job graphs (right figure). Next, the job model generates the task dependency by the following iterative process. Let the number of predecessors and the number of nodes at level $l$ by $|pred(n_i)|$ and $|l|$, respectively. Then, the number of dependent tasks for node $i$ at level $l$ is computed by $\max(1, \min(\mathcal{N}(\frac{|l-1|}{3}, 0), |l-1|))$. We connect $n_i$ to randomly selected $|pred(n_i)|$ nodes in $l-1$ level.

\section{Description of Reward functions} \label{sec:append:reward_fn_eim}
Two types of dense and sparse reward functions are used to validate the efficacy of EIM technique. The reward functions are described as follows.
\begin{align}
    R_{dense}(clk) &= C_1\cdot |\hat{\mathcal{G}}_{comp}| +C_2\cdot clk \label{eq:d_rew1} \\ 
    R_{dense2}(clk) &= C_1\cdot |\hat{\mathcal{G}}_{comp}| \label{eq:d_rew2} \\
    R_{sparse}(clk) &= 0\cdot\mathds{1}_{[clk<CLK-m]} \nonumber\\ &+ C_1\cdot |\hat{\mathcal{G}}_{comp}| \cdot\mathds{1}_{[clk\geq CLK-m]} \label{eq:sp_rew1} \\
    R_{sparse2}(clk) &= 0\cdot\mathds{1}_{[clk\neq CLK]} \nonumber\\ &+C_1\cdot |\hat{\mathcal{G}}_{comp}|\cdot\mathds{1}_{[clk=CLK]}, \label{eq:sp_rew2}
\end{align}
\noindent where $|\hat{\mathcal{G}}_{comp}|$ is the number of newly completed jobs at $clk$, $CLK$ is the end of simulation length, and $m$ is the number of the lastly completed tasks. $C_1$ and $C_2$ are the weights of job completion bonus and penalty for clock signal, respectively. We set +50 for $C_1$ and -0.5 for $C_2$ empirically. All reward functions are computed per simulation clock signal.

\bibliographystyle{plain}
\bibliography{main}

\begin{IEEEbiography}[{\includegraphics[width=1in,height=1.25in,clip,keepaspectratio]{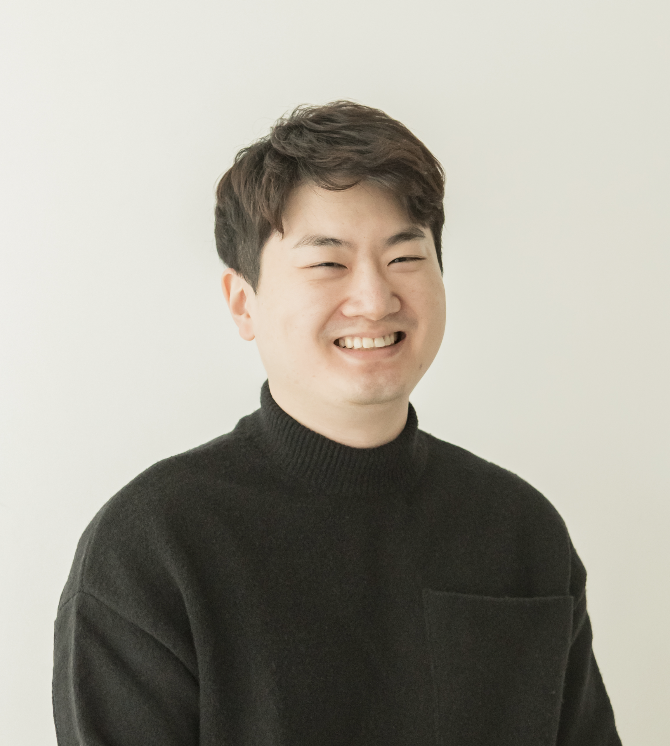}}]{Tegg Taekyong Sung}
Dr. Sung is an expert in the field of deep reinforcement learning with over three years of research and industrial experience. Currently, he is working as a senior research scientist at EpiSys Science, Inc. In the past, he has worked as a research assistant at Kwangwoon University and a visiting researcher at Electronics and Telecommunications Research Institute, South Korea. Dr. Sung’s research interest includes real-world reinforcement learning, machine learning, and graph neural networks. He has authored and co-authored more than 10 publications. He received his Ph.D. from Kwangwoon University, South Korea.
\end{IEEEbiography}

\begin{IEEEbiography}[{\includegraphics[width=1in,height=1.25in,clip,keepaspectratio]{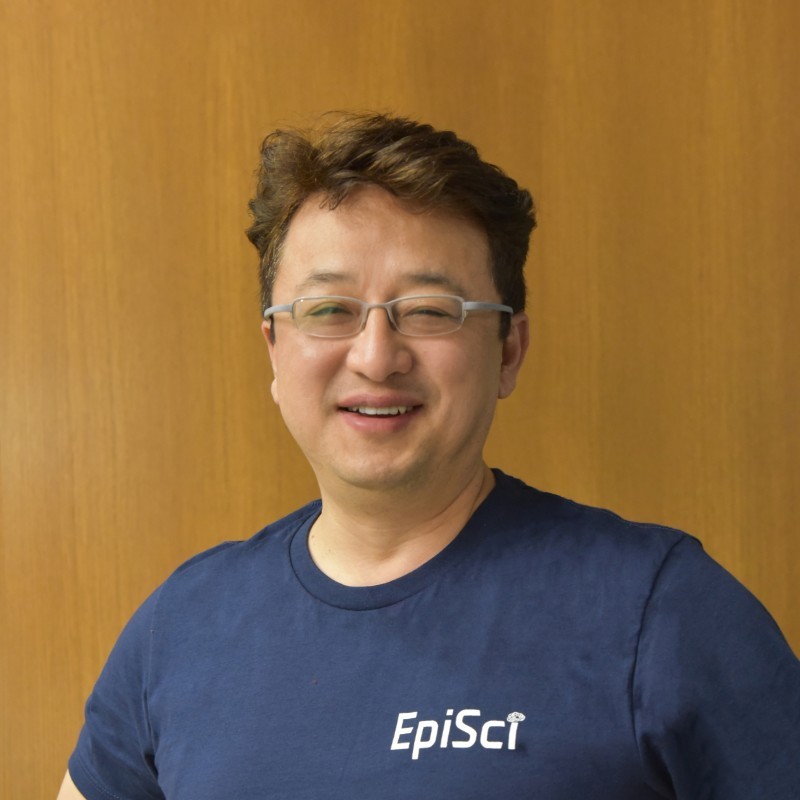}}]{Bo Ryu}
Over the last two decades, Dr. Ryu has accumulated a wealth of successful experiences on high-risk high-pay-off R\&D programs sponsored by DARPA, ONR, AFRL, and various Department of Defense agencies. Before founding EpiSys Science, Inc. in 2013, he served in various technical positions at Hughes, Boeing, San Diego Research Center, and Argon ST. He was responsible for spearheading internal research projects, pursuing new government programs, and performing various government and industry-sponsored projects in the area of self-organizing wireless networking systems. During his time at EpiSci, he has been the PI of over \$20M of R\&D projects from government agencies. He has authored and co-authored more than 40 publications and holds twelve U.S. patents. He received two performance awards for his technical achievements on DARPA’s Adaptive C4ISR Node program and is a recipient of a Meritorious Award from Raytheon in 2001 for technical performance recognition. He received his Ph.D. from Columbia University.
\end{IEEEbiography}

\EOD

\end{document}